%% file: main.tex
\documentclass{article}
\usepackage[preprint, nonatbib]{neurips_2025}

\usepackage{graphicx}
\usepackage{tabularx}

\usepackage[utf8]{inputenc} 
\usepackage[T1]{fontenc}    
\usepackage{hyperref}       
\usepackage{url}            
\usepackage{booktabs}       
\usepackage{amsfonts}       
\usepackage{nicefrac}       
\usepackage{microtype}      
\usepackage{credits}
\usepackage{xcolor}         
\usepackage[
    natbib=true,
    style=numeric,
    sorting=none
]{biblatex}
\usepackage{cleveref}
\usepackage[frozencache,cachedir=.]{minted}
\usepackage{tcolorbox}
\usepackage{fontawesome}
\tcbuselibrary{minted,skins,breakable}
\usepackage{xspace}
\usepackage{longtable}
\usepackage{csquotes}
\usepackage{marvosym}

\newcommand{\chempilelogo}{%
  \raisebox{-0.219\height}{\includegraphics[height=1.25em]{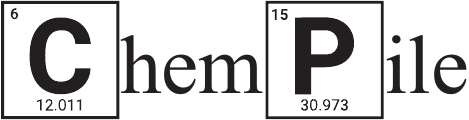}}%
}
\newcommand{\chempileeducationicon}{%
  \raisebox{-0.219\height}{\includegraphics[height=1.25em]{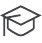}}%
}
\newcommand{\chempilelifticon}{%
  \raisebox{-0.219\height}{\includegraphics[height=1.25em]{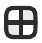}}%
}
\newcommand{\chempilemlifticon}{%
  \raisebox{-0.219\height}{\includegraphics[height=1.25em]{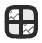}}%
}
\newcommand{\chempilepapericon}{%
  \raisebox{-0.219\height}{\includegraphics[height=1.25em]{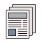}}%
}
\newcommand{\chempilecaptionicon}{%
  \raisebox{-0.219\height}{\includegraphics[height=1.25em]{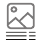}}%
}
\newcommand{\chempilereasoningicon}{%
  \raisebox{-0.219\height}{\includegraphics[height=1.25em]{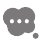}}%
}
\newcommand{\chempilecodeicon}{%
  \raisebox{-0.219\height}{\includegraphics[height=1.25em]{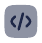}}%
}

\usepackage{stackengine}

\newcommand{\clickableimage}[1]{%
    \raisebox{-0.4ex}{
        \stackinset{c}{0pt}{c}{0pt}{
            \href{#1}{\phantom{\rule{1em}{1em}}}
        }{%
            \includegraphics[width=1em]{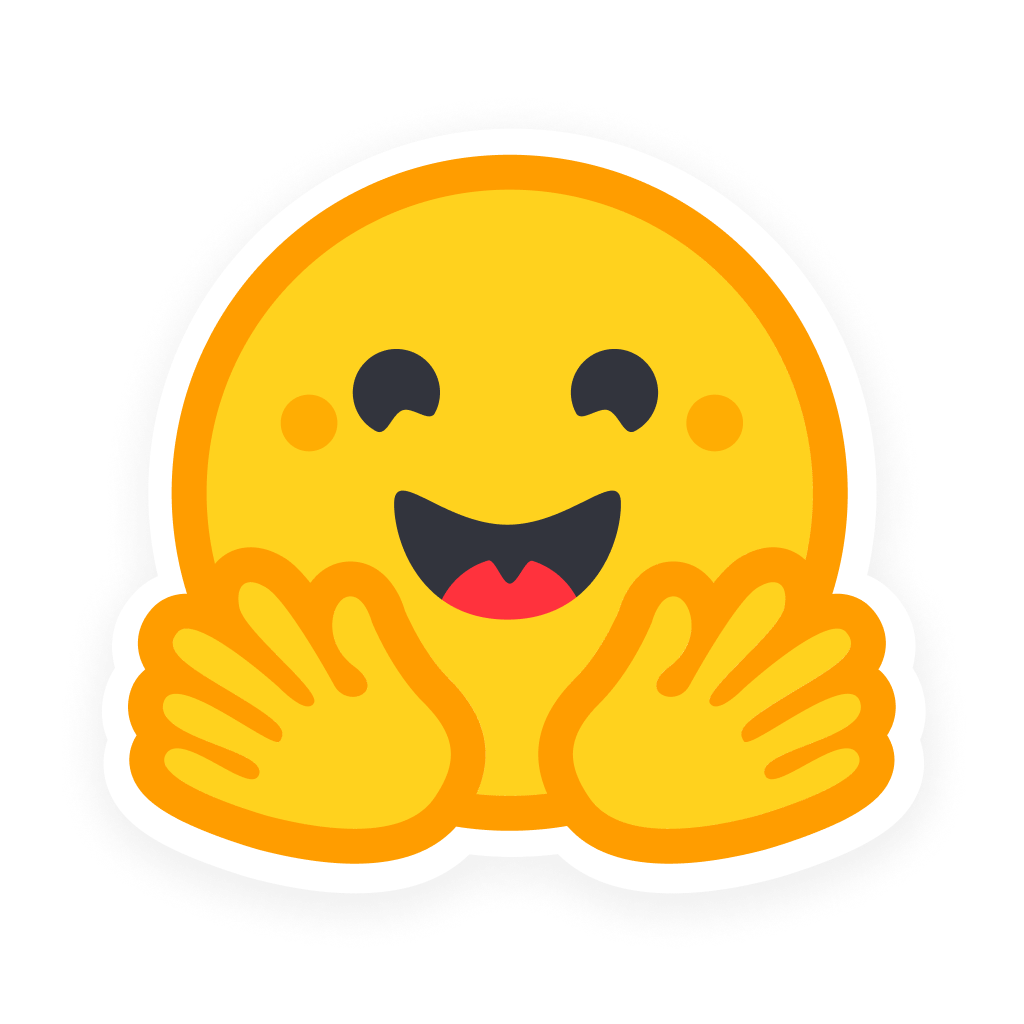}
        }%
    }%
}
\newcommand{\clickableimagegithub}[1]{%
    \raisebox{-0.4ex}{
        \stackinset{c}{0pt}{c}{0pt}{
            \href{#1}{\phantom{\rule{1em}{1em}}}
        }{%
            \includegraphics[width=1em]{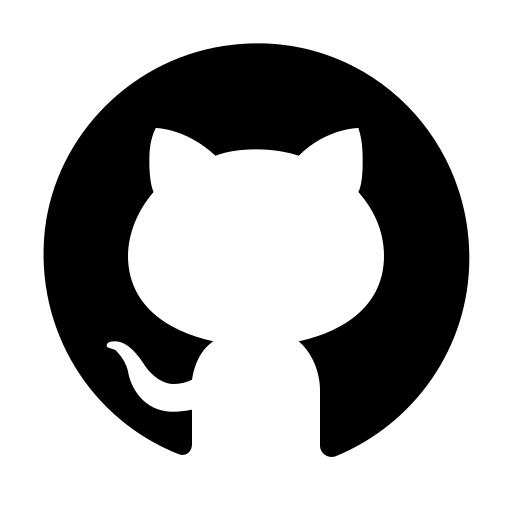}
        }%
    }%
}
\newcolumntype{Y}{>{\centering\arraybackslash}X}

\newcommand{\chempile}{ChemPile\xspace}

\newtcblisting{codebox}[3][]{
  enhanced,
  breakable,
  colback=gray!5,
  colframe=gray!40!black,
  fonttitle=\bfseries,
  coltitle=black,
  attach boxed title to top left={yshift=-2mm, xshift=3mm},
  boxed title style={
    colback=white,
    colframe=gray!40!black,
    sharp corners,
  },
  overlay={
    \node[anchor=north east, font=\scriptsize\sffamily, text=gray!70!black] 
    at ([xshift=-5pt]frame.north east) {\faCode};
  },
  title=#2,
  listing only,
  listing engine=minted,
  minted style=default,
  minted language=python,
  minted options={
    fontsize=\small,
    linenos,
    numbersep=5pt,
    breaklines,
    autogobble,
    frame=none,
    framesep=2mm,
    #1
  },
  label={#3},  
}

\title{\chempilelogo: A 250\,GB Diverse and Curated Dataset for Chemical Foundation Models}

\author{%
  Adrian Mirza~\Letter\\
  HIPOLE Jena \& FSU Jena \\
  \And
  Nawaf Alampara \\
  FSU Jena \\
  \And
  Martiño Ríos-García \\
  FSU Jena \\
  \And
  Mohamed Abdelalim \\
  Independent researcher \\
  \And
  Jack Butler \\
  Faculty \\
  \And
  Bethany Connolly \\ 
  Faculty
  \And
  Tunca Dogan \\
  Hacettepe University \\
  \And
  Marianna Nezhurina \\
  JSC, LAION \\
  \And
  Bünyamin Şen \\
  Hacettepe University
  \And
  Santosh Tirunagari\\
    EMBL-EBI
  \And
   Mark Worrall \\ 
   Faculty\\ 
   \And
   Adamo Young\\
   University of Toronto
  \And
  Philippe Schwaller \\
  LIAC and NCCR Catalysis
  \And
  Michael Pieler~\Letter\\
  Independent researcher \\
  \And
  Kevin Maik Jablonka~\Letter\\
  HIPOLE Jena, FSU Jena, CEEC Jena, JCSM Jena \\
  \And
  \texttt{ \Letter ~ \texttt{andrian.mirza@uni-jena.de}, \texttt{michael.pieler@gmail.com}, \texttt{mail@kjablonka.com}.}
    \And
  \normalfont{Full affiliations are listed in the appendix}.
}

\input{sections/authors.tex}

\begin{document}

\maketitle

\begin{abstract}
Foundation models have shown remarkable success across scientific domains, yet their impact in chemistry remains limited due to the absence of diverse, large-scale, high-quality datasets that reflect the field's multifaceted nature. 
We present the \chempilelogo, an open dataset containing over 75 billion tokens of curated chemical data, specifically built for training and evaluating general-purpose models in the chemical sciences. 
The dataset mirrors the human learning journey through chemistry---from educational foundations to specialized expertise---spanning multiple modalities and content types including structured data in diverse chemical representations (SMILES, SELFIES, IUPAC names, InChI, molecular renderings), scientific and educational text, executable code, and chemical images. \chempile integrates foundational knowledge (textbooks, lecture notes),  specialized expertise (scientific articles and language-interfaced data), visual understanding (molecular structures, diagrams), and advanced reasoning (problem-solving traces and code)---mirroring how human chemists develop expertise through diverse learning materials and experiences. Constructed through hundreds of hours of expert curation, the \chempile captures both foundational concepts and domain-specific complexity. We provide standardized training, validation, and test splits, enabling robust benchmarking. \chempile is openly released via HuggingFace with a consistent API, permissive license, and detailed documentation. We hope the \chempile will serve as a catalyst for chemical AI, enabling the development of the next generation of chemical foundation models.
\end{abstract}

\input{sections/introduction}
\input{sections/related_work}
\input{sections/results}

\input{sections/methods}
\input{sections/conclusions}

\section{Acknowledgments}
This work was supported by the Carl Zeiss Foundation and by Intel and Merck via the AWASES programme.

Parts of A.M.'s work were supported as part of the \enquote{SOL-AI} project funded by the Helmholtz Foundation model initiative.

K.M.J.\ is part of the NFDI consortium FAIRmat funded by the Deutsche Forschungsgemeinschaft (DFG, German Research Foundation) – project 460197019.

P.S.\ acknowledges support from the NCCR Catalysis (grant number 225147), a National Centre of Competence in Research funded by the Swiss National Science Foundation.

M.N.\ acknowledges funding by the Federal Ministry of Education and Research of Germany (BMBF) under grant no. 01IS22094B (WestAI - AI Service Center West), under grant no. 01IS24085C (OPENHAFM) and under the grant 16HPC117K (MINERVA), as well as co-funding by EU from EuroHPC Joint Undertaking programm under grant no. 101182737 (MINERVA) and from Digital Europe Programme under grant no. 101195233 (openEuroLLM).

In addition, we thank the OpenBioML.org community and their ChemNLP project team as well as Prof.\ Andrew White (FutureHouse and University of Rochester, US) and Prof.\ David Windridge (Middlesex University, UK) for valuable discussions.
We also thank Stability.AI for the access to its HPC cluster. 

We thank Anagha Aneesh, Mara Schilling-Wilhelmi, and Meiling Sun for feedback on the manuscript.

\printbibliography

\clearpage
\appendix
\input{sections/appendix}

\end{document}

%% file: sections/authors.tex
\credit{Adrian~Mirza}       {1,1,1,0,1,1,1,0,1,0,1,1,1,1}
\credit{Nawaf~Alampara}     {0,1,1,0,1,1,0,0,1,0,1,1,1,1}
\credit{Martiño~Ríos-García}{0,1,1,0,1,1,0,0,1,0,1,0,1,1}

\credit{Mohamed~Abdelalim}  {0,1,0,0,0,0,0,0,0,0,0,0,0,1}
\credit{Jack~Butler}        {0,1,0,0,0,0,0,0,0,0,0,0,0,1}
\credit{Bethany~Connolly}   {0,1,0,0,0,0,0,0,0,0,0,0,0,1}
\credit{Tunca~Dogan}        {0,1,0,0,0,0,0,0,0,0,0,0,0,1}
\credit{Marianna~Nezhurina} {0,1,0,0,0,0,0,0,0,0,0,0,0,1}
\credit{Bünyamin~Şen}       {0,1,0,0,0,0,0,0,0,0,0,0,0,1}
\credit{Santosh~Tirunagari} {0,1,0,0,0,0,0,0,0,0,0,0,0,1}
\credit{Mark~Worrall}       {0,1,0,0,0,0,0,0,0,0,0,0,0,1}
\credit{Adamo~Young}        {0,1,0,0,0,0,0,0,0,0,0,0,0,1}

\credit{Philippe~Schwaller} {0,1,0,1,0,0,0,1,0,0,0,0,0,1}
\credit{Michael~Pieler}     {1,1,0,1,1,1,1,1,1,1,1,0,0,1}
\credit{Kevin~Maik~Jablonka}{1,1,0,1,1,1,1,1,1,1,1,0,1,1}

%% file: sections/introduction.tex
\section{Introduction}
Foundation models are transforming science, with particularly promising applications in the chemical sciences~\cite{white2023future, Ramos_2025, Jablonka_2023}. 
Progress in this field could fundamentally advance drug discovery, accelerate materials development for energy transition, and provide new solutions for climate change mitigation~\cite{Yao_2022}. 
The potential societal impact is immense.
Recent developments demonstrate that large language models (LLMs) can already answer chemical queries~\cite{mirza2024large, skarlinski2024language, Buehler_2024mechgpt}, predict molecular properties~\cite{qian2023large, jablonka2024leveraging, zhong2024benchmarking, rubungo2023llm0prop0, liu2024moleculargpt0} or crystal structures~\cite{Antunes_2024, gan2025large, gruver2024fine0tuned},  and direct experiments~\cite{Wei_2025, M_Bran_2024, Darvish_2025, boiko2023autonomous}. 
Yet, their performance is often brittle, with shallow reasoning and poor generalization beyond narrow domains~\cite{Binz_2025, Buehler_2024, boyko2023interdisciplinary}. 
These limitations likely stem not (only) from architectural constraints, but from the data on which these models are trained.
Current chemical datasets are fragmented and narrowly focused. Most are confined to a single modality---such as SMILES strings~\cite{weininger1988smiles}---and few capture the underlying reasoning or contextual knowledge that defines chemical understanding. 
Moreover, they are seldom curated with machine learning in mind, leading to issues with inconsistency, data leakage, and poor coverage of fundamental principles \cite{alampara2024probing}.
As a result, existing foundation models in chemistry struggle to learn generalizable patterns, reason across domains, or provide interpretable outputs. \\
To address this, we introduce  \chempilelogo, a large, multimodal open dataset designed to support the training and evaluation of foundation models in chemistry. 
The \chempile is the result of an extensive, community-driven effort, involving hundreds of hours of expert curation, cleaning, and annotation. It provides a unified interface for diverse, multimodal data and is built to serve as a foundational resource for chemical foundation models.
The \chempile mirrors the journey of chemical expertise development in humans---from foundational concepts to specialized knowledge to advanced reasoning---through its diverse content types collected in different subsets:

\begin{itemize}
\item \textbf{\chempilelogo-Education:} Captures foundational core knowledge through curated educational content---similar to how students build conceptual understanding through textbooks and lectures.

\item \textbf{\chempilelogo-Paper:} Incorporates curated scientific literature filtered for chemical content---allowing the models to learn from the frontiers of science.

\item \textbf{\chempilelogo-(m)LIFT:} Provides structured factual knowledge through language-interfaced~\cite{dinh2022lift0} tabular datasets with chemical information in multiple representations (IUPAC, SELFIES, InChI, images) --- allowing the model to learn nuanced structure-property-function relationships.

\item \textbf{\chempilelogo-Reasoning:} Compiles explicit reasoning traces for chemical problems --- to allow models to learn reasoning which is needed to solve advanced chemical problems.

\item \textbf{\chempilelogo-Code:} Includes chemical code ---reflecting that code has often been shown to increase model capabilities~\cite{DBLP:conf/iclr/MaL0Z0W024,albalak2024survey}.

\item \textbf{\chempilelogo-Caption:} Compiles pairs of chemical images with the corresponding descriptive text---reflecting that chemical information is typically multimodal and requires joint reasoning over different modalities such as images or text. 
\end{itemize}

Just as human chemists learn through diverse materials and experiences---textbooks for foundations, laboratory work for hands-on skills, research papers for specialized knowledge, and problem-solving for developing reasoning---\chempile's varied content types aim to provide a comprehensive learning environment for chemical AI.

The core features of the \chempile are:
\begin{itemize}
\item \textbf{Scale:} To our knowledge, \chempile is the largest open curated chemical corpus, providing sufficient data volume for foundation model training and scaling studies.

\item \textbf{Expert curation:} The \chempile has been rigorously cleaned, annotated, and reviewed by domain experts through an extensive collaborative effort. 

\item \textbf{Content type diversity:} The \chempile combines different kinds of content on a spectrum from conceptual understanding (\chempile-Education), over detailed knowledge (\chempile-(M)LIFT), to advanced multimodal reasoning (\chempile-Reasoning, \chempile-Code, \chempile-Caption), covering materials that mirror the human chemist's educational journey.

\item \textbf{Chemical diversity:} The \chempile spans the entire spectrum from biochemistry to materials science, enabling research on domain adaptation and knowledge transfer across chemical subfields that were previously siloed.

\item \textbf{Multimodality:} The \chempile integrates images with captions, molecular and crystal representations in various formats, chemical drawings, and other visual elements essential to chemical communication, providing a foundation for multimodal models.

\item \textbf{Ease of use:} The \chempile is hosted on HuggingFace under a consistent API for public access with a permissive license. We provide recommended training/validation/test splits based on analysis of chemical compounds and extensive documentation (\href{https://chempile.lamalab.org}{chempile.lamalab.org}) to facilitate immediate research use.
\end{itemize}

By centralizing high-quality chemical data in a machine learning-ready format that reflects the multifaceted nature of chemical expertise, we hope that the \chempile will catalyze innovation at the intersection of AI and chemistry. 
The \chempile aims to not just be a dataset, but a bridge between disciplines that will enable a new generation of researchers to contribute to chemical AI and accelerate scientific discovery.

\begin{figure}
\centering
\includegraphics[width=\textwidth]{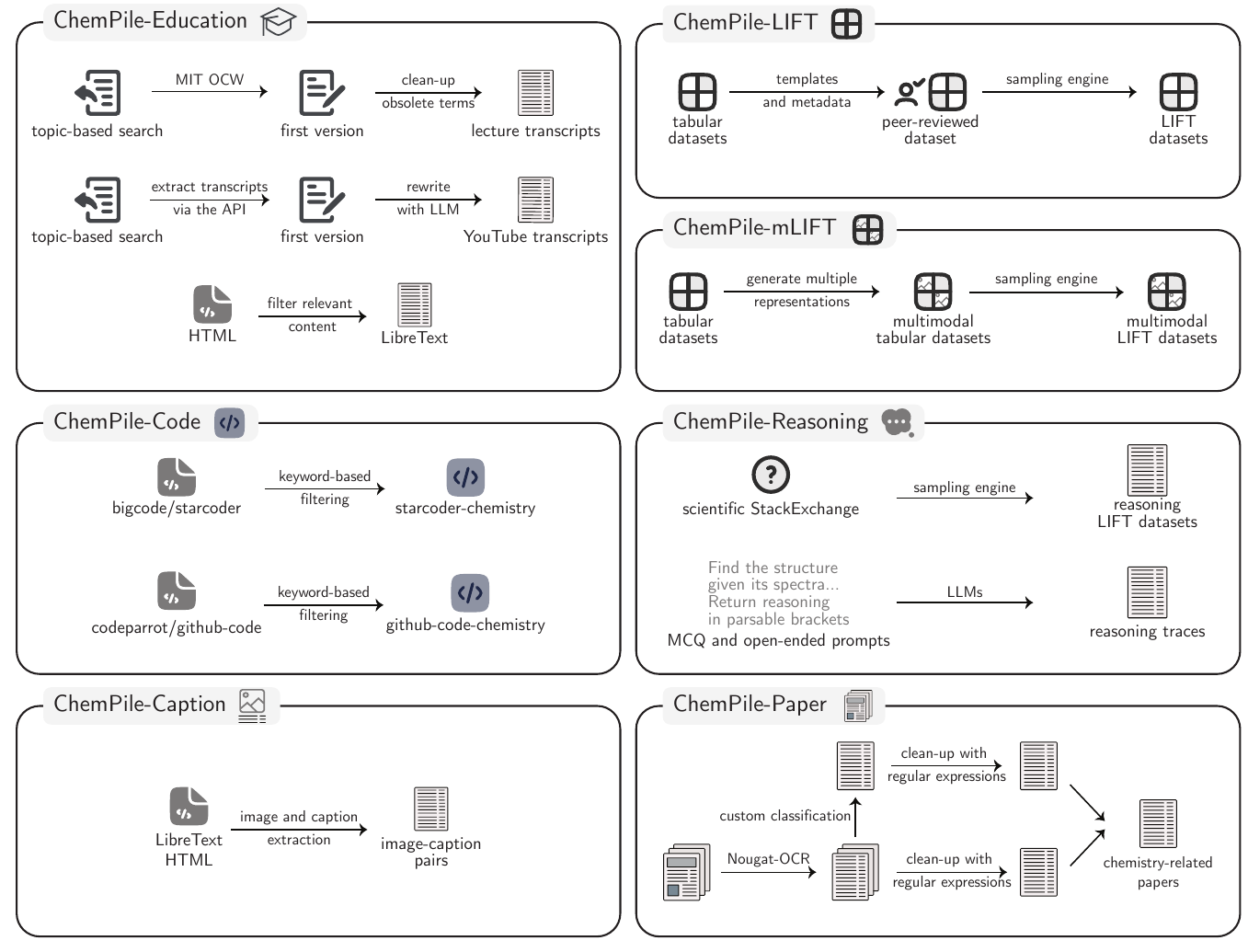}
\caption{\textbf{Overview of the \chempile and its curation process.} The figure illustrates the dataset creation process. Education and Caption consist of gathering resources from online resources. Code and (m)LIFT are based on dataset content, for the first filtering from general datasets, while for the second, by filling templates with the data. For \chempile-Paper, the content is collected by filtering and processing published open-source papers. Finally, reasoning is based on distilling knowledge from LLMs and processing data from Stack Exchange. The resulting datasets are published in a format that is very easy to use on HuggingFace.}
\label{fig}
\end{figure}

%% file: sections/related_work.tex
\section{Related work}
The \chempile is the first dataset that combines diverse content types and chemical subdisciplines under a consistent interface, addressing several key limitations in existing resources. To achieve this, \chempile builds on a foundation of prior work.

\subsection{Datasets for training of foundation models}
Web-scale data has become the standard approach for pre-training foundation models~\cite{gao2020pile, raffel2020exploring}, with empirical scaling laws suggesting that performance improves with dataset size~\cite{hoffmann2022training, kaplan2020scaling,hoffmann2022training}. However, recent studies challenge the \enquote{more data is always better} paradigm, exploring data-effective learning approaches that focus on quality and representativeness rather than sheer volume~\cite{marion2023less, gunasekar2023textbooks, gadre2023datacomp, DBLP:conf/nips/PenedoKALMRW024}.
Current state-of-the-art training pipelines use carefully constructed mixtures of different data types including natural text, code, textbooks, and reasoning traces to improve model capabilities~\cite{soldaini2024dolma,albalak2024survey, feng2024maximize}. 
While this trend toward high-quality, diverse data mixtures has transformed general-purpose AI, the chemical domain has not yet benefited from similar approaches.

For multimodal foundation models, image-text pairs represent the primary training data format, typically sourced from web pages containing images with associated alt-text, captions, or surrounding text~\cite{laion_5b}. 
However, equivalent multimodal resources have been largely absent in the chemical domain until the \chempile.

\subsection{Chemical datasets}
Traditional chemistry datasets have largely relied on tabular formats as compiled in MoleculeNet~\cite{wu2018moleculenet} or Therapeutic Data Commons~\cite{tdc}. 
Resources like PubChem~\cite{Kim_2015} and UniProt~\cite{uniprot2019uniprot} provide large collections of molecular structures or protein sequences for tasks in (bio)molecular property prediction. 
A critical limitation is that these resources cannot be directly used for training LLMs as they require conversion into natural language through templates that demand significant domain knowledge to set up properly~\cite{gonzales2024evaluating}.

While experimentally derived datasets are typically small or medium-sized, larger resources such as QM9~\cite{Ramakrishnan_2014} have been compiled based on computational screenings. 
However, these datasets may not fully capture real-world variations and experimental noise. Other resources such as the USPTO database~\cite{Lowe2017} are, for example, patent-derived and come with corresponding biases~\cite{Schneider_2016, Jia_2019, Raccuglia_2016}.

A fundamental challenge remains the integration of information from different sources, chemical subfields, and modalities. 
Scientific information is frequently distributed across multiple datasets, making it difficult to assemble comprehensive resources that reflect the true complexity of chemical phenomena~\cite{Ongari_2022}. 
The \chempile explicitly addresses this fragmentation by unifying diverse chemical information under a consistent interface.

In addition, it is important to realize that molecules can be represented in various string formats, including IUPAC names, SMILES, DeepSMILES, SELFIES, and InChI~\cite{Krenn_2022}. 
Currently, there is no consensus on which representation is optimal for training chemical foundation models. 
To enable the systematic comparison of their effectiveness, the \chempile includes multiple representations for the same molecules.

\subsection{Chemical text and multimodal datasets}
Recent efforts to create specialized chemical datasets include the Mol-instructions dataset~\cite{fang2023mol0instructions0}, which provides around 2 million biomolecular and protein-related instructions. 
In the multimodal space, several specialized resources such as MoMu~\cite{su2022molecular}, PubChemSTM~\cite{Liu_2023}, Llamole~\cite{liu2024multimodal},  and MultiMat~\cite{moro2023multimodal} have emerged.

While these specialized datasets represent important advances, they remain limited in scope and typically focus on a single chemical subdomain or modality pairing. 

The \chempile builds upon these efforts by providing a unified resource that spans multiple chemical subfields and integrates all relevant modalities under a consistent framework, addressing the fragmentation, narrow focus, modality restrictions, and inconsistent formats of existing chemical datasets.

%% file: sections/results.tex
\section{Overview of the \chempile}
The \chempile is distinct in scale, breadth, curation quality, and ease-of-use.

\begin{figure}[ht]
    \centering
    \includegraphics[width=\textwidth]{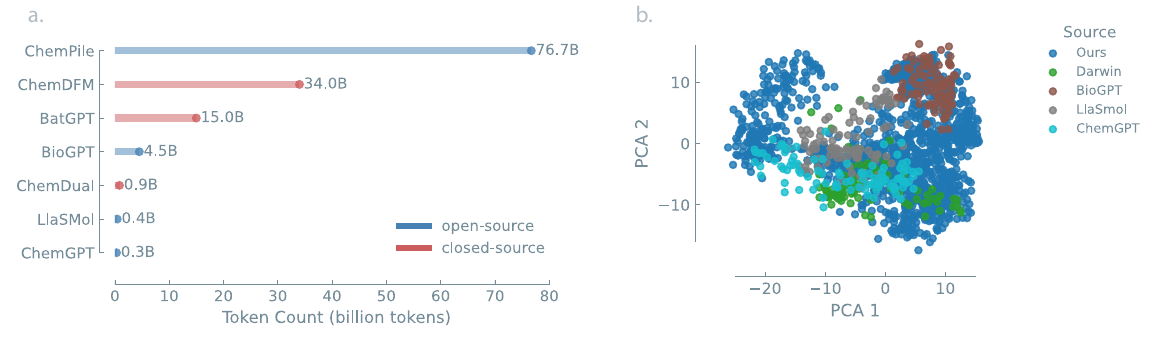}
    \caption{\textbf{(a): Token count comparison between the \chempile dataset and other domain-specific large datasets used to train foundation models.} \texttt{ChemDFM}~\cite{zhao2024chemdfm} is a foundation model for chemistry trained on 34B tokens in chemistry-related papers and textbooks augmented with general text (49M tokens), BatGPT~\cite{yang2024batgpt} is a foundation model for chemical engineering, BioGPT~\cite{luo2022biogpt} for biology, and ChemGPT~\cite{frey2023neural} is a foundation model trained only on SMILES string. LlaSMol~\cite{yu2024llasmol0} is an instruction-tuning dataset for chemistry. ChemDual~\cite{lin2025enhancing} is a 4.4 million instruction dataset for chemical reactions. The value for BioGPT is an estimate based on: 15 M abstracts × 250 words × 1.2 tokens $\approx$ 4.5 B tokens. We compute the estimate for LlaSMol based on the published HuggingFace dataset. The scale of our dataset exceeds any of the corpora used to pre-train or fine-tune existing chemistry LLMs. 
    \textbf{(b): Embedded datapoints sampled from various subsets of ChemPile vs other public datasets}. Note, only the instruction tuning data made public by the authors of Darwin \cite{xie2024darwin} is used. We embed only the first 512 tokens of each sampled document using the \texttt{specter2-base} model provided by \textcite{Singh2022SciRepEvalAM}. Along PCA, we provide UMAP and TSNE plots in \Cref{app:tsne_umap}.}
    \label{fig:scale}
\end{figure}

\paragraph{Scale}
One of the most essential characteristics of a dataset for training foundation models is its scale~\cite{kaplan2020scaling}.
For reference, we compare \chempile to other domain-specific foundation models (\Cref{fig:scale}). 
\texttt{ChemDFM}~\cite{zhao2024chemdfm} is the largest chemical foundation model that has been reported. It has been trained on a dataset of 34B tokens which, however, has not been released. 
Even though it contains general-purpose data (such as Wikipedia and the WuDao Corpora~\cite{yuan2021wudaocorpora}), it is still more than 50\% smaller than the \chempile.
Other notable chemistry datasets, such as LlaSMol~\cite{yu2024llasmol0} and ChemDual~\cite{lin2025enhancing}, are orders of magnitude smaller.

As \Cref{fig:scale}a) illustrates, \chempile is the largest open chemical dataset we are aware of and the only one that reaches a scale that is meaningful for training foundation models.

\paragraph{Diversity} 
The \chempile is not only large but also diverse. Data mixing for training LLMs is still not fully understood and is an active field of research. Different mixes typically yield different generalization performance~\cite{soldaini2024dolma,rae2021scaling,ye2024data}. To enable such research, the \chempile was designed to be maximally diverse. \Cref{fig:scale}b) illustrates this. In this figure, we showcase that the embeddings of data from the \chempile span a larger space than data from many other chemical datasets combined.

We achieve this in multiple ways: First, sampling and curating data from very different sources and, second, by representing chemical entities in various modalities and text forms.

In contrast to other large chemical datasets, \chempile is a systematic collection of multiple subsets that were curated to encompass specific knowledge or to potentially convey specific abilities to models trained on those subsets. These subsets, which we describe in detail in \Cref{sec:chempile_details}, contain data sampled from very different sources such as structured chemical datasets, recordings of lectures, or data we created from scratch.  

\begin{figure}[h]
    \centering
    \includegraphics[width=\textwidth]{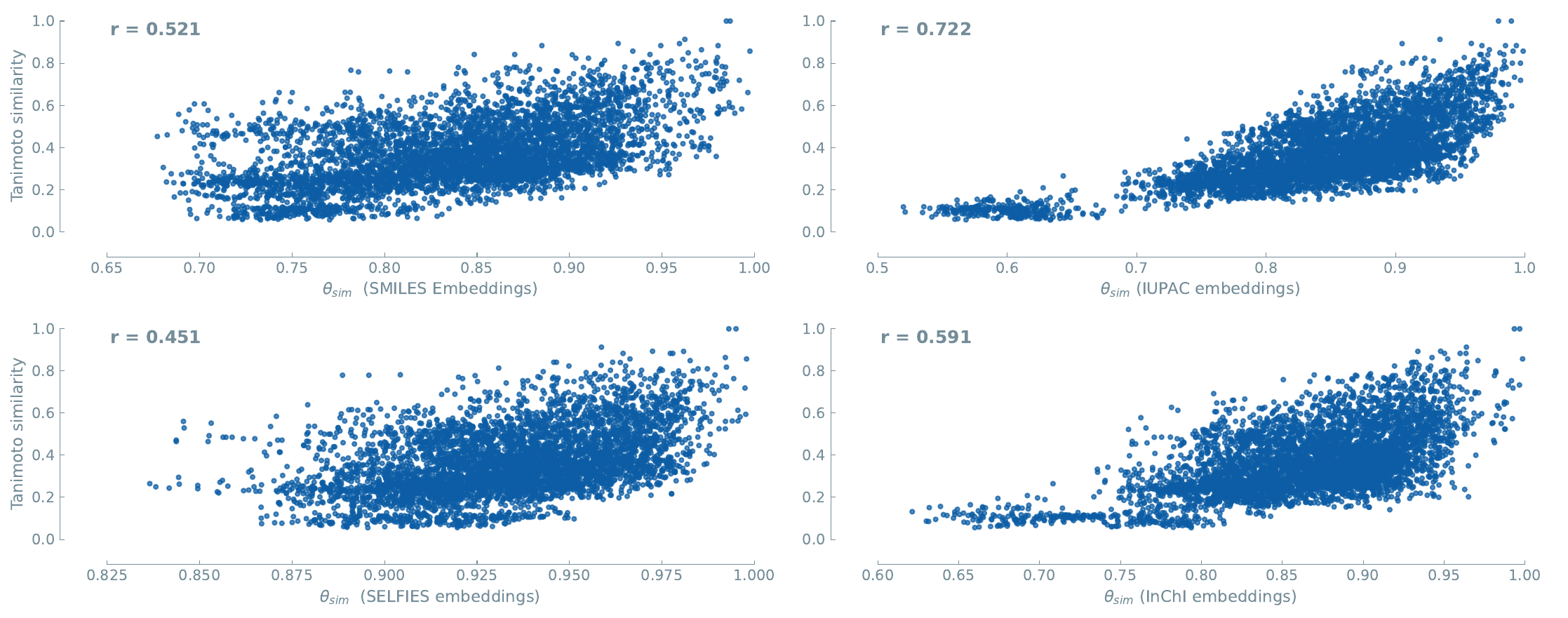}
    \caption{\textbf{Correlation between the Tanimoto similarity and the cosine similarity ($\theta_{sim}$) of the embeddings for most common chemical representations}. The correlation is shown for four representation embeddings: SMILES (top left), IUPAC name (top right), SELFIES (bottom left), and InChI (bottom right).  For the four subplots, we show the Pearson correlation \textbf{r} in the top left corner of all subplots.}
    \label{fig:smiles_iupac_correlations}
\end{figure}

In addition, \chempile considers the fact that chemical entities, such as molecules, can be represented in diverse forms. This includes diverse string representations, such as IUPAC names, SELFIES~\cite{krenn2020self}, SMILES~\cite{weininger1988smiles}, and InChI~\cite{heller2015inchi}, but also molecular drawings~\cite{brinkhaus2022randepict} in addition to images from chemical textbooks.\\
This feature of the \chempile is relevant because while SMILES are widely used in cheminformatics, it is not obvious that they are also the best choice for building foundation models. First, SMILES and other chemical representations are not handled in optimally in conventional pretrained tokenizers~\cite{alampara2024mattext, chithrananda2020chemberta0}. This is particularly interesting for finetuning and continued pretraining studies and can be seen in a correlation analysis (\Cref{fig:smiles_iupac_correlations}). If different chemical representations are embedded with existing embedding models, similarity between embeddings of IUPAC names correlates much strongly with established similarity measures---such as the Tanimoto similarity between molecular fingerprints~\cite{bajusz2015tanimoto}---than the embeddings of other molecular representations.

In addition, one might expect benefits from the inclusion of IUPAC names as they are not only closer to common English text but, in particular, the text seen in chemical papers and hence might improve training dynamics.
 
\paragraph{Quality}

Curation quality distinguishes \chempile from previous chemical datasets. Domain specialists manually reviewed each subset, ensuring scientific accuracy and relevance. For \chempile-(m)LIFT, we implemented a systematic verification protocol where chemical experts checked template design, property assignments, and molecular representations. All datasets underwent multiple validation passes to eliminate inconsistencies, incorrect nomenclature, and formatting errors. This curation process, representing hundreds of expert hours, delivers a dataset that captures both foundational concepts and specialized knowledge with high fidelity.

\paragraph{Ease of use}

The \chempile prioritizes accessibility for researchers from different domains through consistent interfaces across all datasets. 
We host the entire data collection on HuggingFace with uniform formatting and comprehensive documentation (\href{https://chempile.lamalab.org}{chempile.lamalab.org}). Each subset includes detailed metadata, usage examples, and explicit training/validation/test splits designed to prevent chemical structure leakage between partitions. 
The modular architecture allows researchers to use specific subsets independently (\Cref{code:chempile-caption}), or combine them as needed (\Cref{code:chempile-pretraining}).
This accessibility reduces barriers to entry for researchers from both machine learning and chemistry backgrounds, enabling immediate application to foundation model training, specialized fine-tuning, or directed research on particular chemical domains.

%% file: sections/methods.tex
\section{Diving into the \chempile}
\label{sec:chempile_details}
The \chempile can be used for many tasks, but the focus is on the training of general-purpose foundation models for the chemical sciences. This section provides detailed information on the seven datasets, detailed in \Cref{tab:token-count}, making up the ChemPile data and their curation process.

\begin{table}[!h]
    \centering
        \caption{\textbf{Token count and size in GB of the datasets}. The number of tokens was estimated using \texttt{tiktoken}~\cite{tiktoken} with the model \texttt{gpt2}. The dataset size indicated in this table corresponds to the compressed file size following conversion to the Parquet format. Note that the number of figures for the multimodal datasets and the number of images is equal to the number of documents---one image per entry.}
    \begin{tabularx}{\textwidth}{lYYYY}
        \toprule
        Dataset & Size (GB) &  Number of text tokens & Number of documents & HuggingFace dataset \\
        \midrule
        \chempilelogo-Education\chempileeducationicon & 0,25 & 130M  & 66,9K & \clickableimage{https://huggingface.co/datasets/jablonkagroup/chempile-education} \\
        \chempilelogo-Paper\chempilepapericon         & 31,6 & 14,1B & 11,7M & \clickableimage{https://huggingface.co/datasets/jablonkagroup/chempile-paper} \\
        \chempilelogo-LIFT\chempilelifticon           & 49.1 & 29,4B & 185M  & \clickableimage{https://huggingface.co/datasets/jablonkagroup/chempile-lift-merged} \\
        \chempilelogo-mLIFT\chempilemlifticon         & 155  & 15,0B & 61.6M & \clickableimage{https://huggingface.co/datasets/jablonkagroup/chempile-mlift} \\
        \chempilelogo-Code\chempilecodeicon           & 15,6 & 18,0B & 2,27M & \clickableimage{https://huggingface.co/datasets/jablonkagroup/chempile-code} \\
        \chempilelogo-Reasoning\chempilereasoningicon & 0,10 & 20,0M & 72,9K & \clickableimage{https://huggingface.co/datasets/jablonkagroup/chempile-reasoning} \\
        \chempilelogo-Caption\chempilecaptionicon     & 3,23 & 10,3M & 100K  & \clickableimage{https://huggingface.co/datasets/jablonkagroup/chempile-caption} \\
        \midrule
        \chempilelogo                                 & 255 & 76,7B  & 260M  & \clickableimage{https://huggingface.co/collections/jablonkagroup/the-chempile-6824e88c60d3286ba9b0dae1} \\
        \bottomrule
    \end{tabularx}
    \label{tab:token-count}
\end{table}

\subsection{\chempile-Education}

\chempile-Education contains (foundational) knowledge exposition from lectures and textbooks as well as worked practice problems (see \Cref{fig:chempile_smol}).

The data collection involved four distinct methodologies tailored to source-specific characteristics.  
\textbf{LibreTexts Chemistry} contains open-source chemistry textbooks, which we mined using a pipeline that systematically parses HTML documents, stripping non-content elements to compile a chemically focused corpus of 114 million tokens. 
\textbf{MIT OpenCourseWare} lecture materials were programmatically retrieved through topic-specific searches (biology, chemistry, chemical engineering, physics), with course names and download links preserved. \textbf{YouTube course transcripts} were sourced via LLM-generated keyword queries, restricted to Creative Commons-licensed videos, and refined using \texttt{GPT-4.1} to correct transcription errors and enhance coherence. 
\textbf{US Olympiad problems} from 2003 to 2024 were manually processed using the \texttt{Gemini 2.0 Flash} model, which aligned PDF-based questions and solutions into JSON-structured metadata and selected solutions exceeding 250 characters. 

A detailed explanation of the workflow for each of the sources can be found in \Cref{sec:app-education}

\subsection{\chempile-Paper}
As a resource for cutting-edge applications of chemical knowledge and reasoning, we also curated a dataset of papers from diverse repositories in \chempile-Paper.
The \textbf{EuroPMC} dataset~\cite{rosonovski2024europe}, comprising 27 million abstracts and 5 million full-text articles, was filtered using a BERT-based multilabel classifier trained on the CAMEL datasets (20,000 examples per discipline) \cite{li2023camel} and validated against FineWebMath \cite{allal2025smollm2smolgoesbig} annotations (F$_1$-score $\approx$ 0.77 on 150 entries we manually annotated). Chemistry-related content was identified by analyzing the first five 512-token chunks per document with 50-token overlaps, yielding 3.3 billion tokens. 
Preprints from \textbf{ChemRxiv}, \textbf{BioRxiv}, and \textbf{MedRxiv} were collected via PaperScraper~\cite{born2021trends}, processed with Nougat~\cite{blecher2023nougat0} for text extraction, and enriched with metadata (license, publication date, authors, title). 
\textbf{ArXiv} submissions were filtered by materials science and physical chemistry keywords (e.g., \texttt{cond-mat.mtrl-sci}), with PDFs retrieved via PaperScraper. For all scientific articles, we employed a postprocessing pipeline that removed text that is not directly linked to chemical information (e.g., authors, acknowledgments, page numbers) as explained in \Cref{sec:clean_papers}.
Additionally, we included \textbf{Materials Safety Data Sheets (MSDS)} as structured tabular data (H/P statements) and natural text, ensuring comprehensive coverage of safety information. This multi-source approach balances breadth and domain specificity across literature, preprints, and regulatory documents. 
A more concise explanation about the sources and in the data-curation procedure is in \Cref{sec:app-paper}.

\subsection{\chempile-(m)LIFT}
In \chempile-(m)LIFT, we compile language-interfaced tabular data about properties of molecules, materials, and reactions to allow models to learn intricate structure-property-function relationships. 

\paragraph{Curation process} We manually collected and annotated structured chemical datasets. In the annotation process, domain experts not only annotated the meaning (in many cases including links to ontologies) and possible namings of columns but also created multiple templates that use the tabular data in different language-interfaced settings such as (multiobjective) property prediction or inverse design. The entire curation process was organized via Pull Requests on GitHub which were reviewed by at least one other domain expert.
We provide examples of some of those templates in \Cref{tab:templates} (a total of 1636 templates have been manually curated). All curation scripts and metadata files are available on GitHub\clickableimagegithub{https://github.com/lamalab-org/chempile/tree/main/data/tabular}.

\paragraph{Sampling engine}

The language-interfaced tabular data has been generated with a sampling engine that includes several functionalities: flexible multiple-choice question generation (including permutation of enumeration symbols),  synonym sampling (e.g., diverse sampling of property names or molecular representations), as well as conditional formatting (e.g., for negations). 
An illustrated example and a more detailed explanation can be found in \Cref{app:sampling}.
In the sampled datasets, we distinguish between completion and instruction type templates and allow the user to select data formatted in specific templates to allow systematic ablation studies~\cite{gonzales2024evaluating}.

\paragraph{\chempile-mLIFT}
Since our annotation process clearly identified columns containing molecular, material, or reaction information, we could systematically compute alternative representations such as SMILES, InChI, SELFIES, IUPAC names, and images for all entries in \chempile-LIFT using cheminformatics tools. In particular, images were created in various styles using a pipeline based on RanDepict~\cite{brinkhaus2022randepict}. 

Interestingly, the generation of IUPAC names at scale is challenging due to the lack of open-source tools that can create IUPAC names based on SMILES. However, the validation can be robustly performed using the open-source IUPAC-to-SMILES converter OPSIN~\cite{lowe2011chemical}. 
Thus, we trained a SMILES-to-IUPAC model based on an encoder-decoder architecture and automatically verified the validity of the outputs of the model using OPSIN.

\subsection{\chempile-Code}
Programming is a crucial part of chemistry research, for example, as part of data analysis or computational chemistry. Hence, it is important to cover chemistry-related code knowledge during training. Moreover, it has been shown that including code datasets during pretraining can improve reasoning~\cite{DBLP:conf/iclr/MaL0Z0W024, aryabumi2024code1}.

To create the \chempile-Code subset, we filter some of the biggest and widely used datasets. 
We use regular-expression-based filters to relevant code snippets pertaining to chemistry, materials science, and biology, as well as specific scientific software packages. 
The majority of the code after filtering is related to simulations, see \Cref{fig:keyword-distribution} (also see the keywords used for filtering \Cref{tab:keywords}). 

The collection primarily comprises a filtered version of the StarCoder and CodeParrot datasets. \textbf{StarCoder}~\cite{li2023starcoder} is a filtered version of the Stack dataset~\cite{kocetkov2022stack1}. \textbf{Codeparrot} is a subset of GitHub-code.
The datasets were deduplicated based on exact hash matching. Furthermore, entries from CodeParrot were removed from Starcoder again based on hash string matching to avoid any overlaps between the two datasets.

\subsection{\chempile-Reasoning}

As training on worked examples and reasoning chains is known to improve the performance of foundation models, we specifically created such datasets.

\chempile-Reasoning combines data from two primary sources. For the first sources, we gathered and filtered content from the \textbf{Chemistry}, \textbf{Matter Modeling}, and \textbf{Physics Stack Exchange} forums. 
The collected data was processed using templates incorporating questions and answers in distinct templates and linguistic styles to enhance diversity. This approach yielded datasets of 12 million, 7 million, and 1.7 million tokens for physics, chemistry, and matter modeling, respectively.

The second source involves \textbf{synthetic reasoning traces} generated by the \texttt{Claude-3.5-Sonnet} and \texttt{Deepseek-R1} models~\cite{deepseek-ai2025deepseek0r10}. These models were prompted to perform spectral elucidation tasks, analyzing molecular spectra to identify corresponding molecules. Over 2 million tokens of distilled synthetic reasoning data were collected through this process. We provide additional methodological details, including data parsing and curation steps, in \Cref{sec:app-reasoning}.

\subsection{\chempile-Caption}

The \chempile-Caption dataset contains over 100,000 text-image pairs focused on foundational chemistry concepts. We sourced images and their corresponding captions and alt texts from \textbf{LibreTexts Chemistry} using HTML parsing. To ensure data quality, we excluded images lacking descriptive text or with fewer than 200 combined characters in their captions and alt texts. This curation process resulted in a high-quality multimodal dataset, as LibreTexts Chemistry content originates from peer-reviewed college courses and textbooks, ensuring reliability and academic relevance.

\subsection{Splits}
Depending on the representation of molecules and macromolecular structures (e.g., proteins or polymers), we split the datasets differently. 
All SMILES across the various tabular datasets are combined into a single list. Then, we apply scaffold splitting (based on the RDKit Murcko scaffolds)~\cite{wu2018moleculenet,bemis1996properties}.
At the same time, we ensure that for all datasets, the validation and test sets are not empty. This ensures the usability of individual language-interfaced tabular datasets for other downstream tasks, such as fine-tuning.

For amino-acid sequences (i.e., proteins), we follow the same procedure for deduplication, but apply random splitting on all sequences across datasets. For datasets without SMILES or amino-acid sequences, we apply random splitting for individual datasets. More details on the splitting procedure are shown in \Cref{app:data_splitting_algo}.

%% file: sections/conclusions.tex
\section{Future work}
\chempile establishes the essential foundation for the next generation of chemical AI, creating a pathway for numerous exciting developments now within reach. 
The infrastructure we've created enables seamless integration of organometallic chemistry datasets, which are currently underrepresented, as specialized representations evolve~\cite{krenn2020self}. 
Our multimodal datasets provide the perfect scaffold for incorporating spectroscopic data, reaction dynamics visualizations, and materials-specific representations. 
The robust splitting methodology in \chempile-(m)LIFT and the \chempile-Paper dataset opens the door to sophisticated chemical entity recognition across papers, a capability that will further enhance model performance through improved deduplication and knowledge integration. 
Our extensible sampling engine can be extended to support data generation for foundation model architectures beyond language models, including GNNs and contrastive models, broadening \chempile's utility.

\section{Conclusions}
The chemical sciences stand at the forefront of AI's potential societal impact, with applications ranging from drug discovery to climate change mitigation. Until now, progress --- for example, in the development of chemical foundation models --- has been constrained by the absence of data resources that reflect chemistry's multifaceted nature. \chempile transforms this landscape by providing the first dataset with meaningful scale and diversity for chemistry. 
By mirroring the human learning journey---from educational foundations to specialized knowledge to multimodal understanding---\chempile creates a comprehensive learning ecosystem for chemical AI.  \chempile serves as a bridge between disciplines that will enable a new generation of researchers to contribute to chemical AI and accelerate scientific discovery.

%% file: sections/appendix.tex
\Large{\textbf{Appendix}}\\
\normalsize
\section{Full affiliations}

\paragraph{FSU Jena}
Laboratory of Organic and Macromolecular Chemistry (IOMC), Friedrich Schiller University Jena, Humboldtstrasse 10, 07743 Jena, Germany
\begin{itemize}
    \item Adrian Mirza
    \item Nawaf Alampara
    \item Martiño Ríos-García
    \item Kevin Maik Jablonka
\end{itemize}
\paragraph{HIPOLE Jena}
Helmholtz Institute for Polymers in Energy Applications Jena (HIPOLE Jena), Lessingstrasse 12-14, 07743 Jena, Germany
\begin{itemize}
    \item Adrian Mirza 
    \item Kevin Maik Jablonka
\end{itemize}

\paragraph{Independent researcher} Mohamed Abdelalim and Michael Pieler

\paragraph{Faculty} Faculty, 160 Old Street, London, UK
\begin{itemize}
    \item Jack Butler 
    \item Bethany Connolly
    \item Mark Worrall
\end{itemize}

\paragraph{Hacettepe University} Biological Data Science Lab, Dept.~of Computer Engineering, Hacettepe University, 06800, Ankara, Türkiye 
 and Dept.~of Health Informatics, Institute of Informatics, Hacettepe University, 06800, Ankara, Türkiye 
 \begin{itemize}
     \item Tunca Dogan
     \item Bünyamin Şen
 \end{itemize}

 \paragraph{JSC} Juelich Supercomputing Center (JSC), Research Center Juelich (FZJ), Germany 
 \begin{itemize}
     \item Marianna Nezhurina
 \end{itemize}

  \paragraph{LAION} 
 \begin{itemize}
     \item Marianna Nezhurina
 \end{itemize}

 \paragraph{EMBL-EBI} Literature Services Team, European Bioinformatics Institute, European Molecular Biology Laboratory (EMBL-EBI), Wellcome Trust Genome Campus, Hinxton, CB10 1SD, Cambridge, United Kingdom. 
 \begin{itemize}
     \item Santosh Tirunagari
 \end{itemize}

\paragraph{University of Toronto} Department of Computer Science, University of Toronto
\begin{itemize}
    \item Adamo Young
\end{itemize}

 \paragraph{LIAC} Laboratory of Artificial Chemical Intelligence (LIAC), Institut des Sciences et Ing\'{e}nierie Chimiques, Ecole Polytechnique F\'{e}d\'{e}rale de Lausanne (EPFL), Lausanne, Switzerland.
\begin{itemize}
    \item Philippe Schwaller
\end{itemize}

\paragraph{NCCR Catalysis} National Centre of Competence in Research (NCCR) Catalysis, Ecole Polytechnique F\'{e}d\'{e}rale de Lausanne (EPFL), Lausanne, Switzerland.
\begin{itemize}
    \item Philippe Schwaller
\end{itemize}

\paragraph{CEEC Jena} Center for Energy and Environmental Chemistry Jena (CEEC Jena), Friedrich Schiller University Jena, Philosophenweg 7a, 07743 Jena, Germany
\begin{itemize}
    \item Kevin Maik Jablonka
\end{itemize}

\paragraph{JCSM Jena} Jena Center for Soft Matter (JCSM), Friedrich Schiller University Jena, Philosophenweg 7, 07743 Jena, Germany
\begin{itemize}
    \item Kevin Maik Jablonka
\end{itemize}

\section{Credits}
The project was conceptualized as part of the ChemNLP project, led by Michael Pieler and Kevin Maik Jablonka. Michael Pieler led the development of the sampling engine, which was refactored by Kevin Maik Jablonka and Adrian Mirza. Postprocessing code for natural text data was developed by Michael Pieler and Kevin Maik Jablonka. Dataset filtering code and models were developed by Nawaf Alampara, who also led the final curation of \chempile-Paper and \chempile-Code. 
Martiño Ríos-García created parts of the \chempile-Education corpus and led the development of the \chempile-website.
Adrian Mirza revised the \chempile-(m)LIFT corpus and led the creation of the HuggingFace collections. 
The final version was compiled by Adrian Mirza, Nawaf Alampara, and Martiño Ríos-García in the research group led by Kevin Maik Jablonka. 
All authors contributed to the data curation. 

 \resizebox{1\linewidth}{!}{\footnotesize\insertcredits}
\normalsize
\input{sections/datasheets}
\clearpage

\section{Licences of the datasets}

The datasets comprising ChemPile operate under heterogeneous licensing agreements reflecting their diverse origins. Specifically, the mLIFT, Education, and Caption datasets are distributed under the Creative Commons Attribution-NonCommercial-ShareAlike 4.0 International License (CC BY-NC-SA 4.0). The Paper dataset employs the more restrictive CC BY-NC-ND 4.0 license, while the Code repository utilizes the Apache 2.0 software license. Notably, the Reasoning dataset features the most permissive terms through its CC BY-SA 4.0 license. This licensing framework preserves the original terms associated with each constituent data source while facilitating transparent reuse guidelines.

\section{Data splitting for tabular datasets}\label{app:data_splitting_algo}

We concatenated all molecules for the datasets containing the SMILES representation. The challenge lied in achieving a consistent train--test--validation split across all such tabular datasets, which are distinct both in terms of the number of molecules, and molecular diversity. We demonstrate the full algorithm to obtain non-empty scaffold splits across tabular datasets as pseudo-code. The full Python implementation can be found on GitHub\clickableimagegithub{https://github.com/lamalab-org/chempile/blob/80a26906988c27bc0659473b6c601a40ab147e3e/data/train_test_split.py}.

\begin{codebox}{Pseudo-code for scaffold splitting across tabular datasets}{code:pseudo-code-splits}
 # STEP 1: Create global split assignments for all molecules
function CreateGlobalMoleculeSplits():
    all_molecules = empty set
    
    # Collect all unique molecules across designated datasets
    for each dataset in scaffold_split_datasets:
        molecules = extract_smiles_from(dataset)
        add molecules to all_molecules
    
    # Convert to list for indexing and shuffle
    all_molecules_list = convert_to_list(all_molecules)
    shuffle(all_molecules_list)
    
    # Assign to splits based on fractions
    train_size = floor(length(all_molecules_list) * train_fraction)
    val_size = floor(length(all_molecules_list) * val_fraction)
    
    train_molecules = all_molecules_list[0 : train_size]
    val_molecules = all_molecules_list[train_size : train_size + val_size]
    test_molecules = all_molecules_list[train_size + val_size : end]
    
    # Save for future reference
    save_to_file("val_molecules.txt", val_molecules)
    save_to_file("test_molecules.txt", test_molecules)
    
    return train_molecules, val_molecules, test_molecules

# STEP 2: Apply consistent splits to all datasets with SMILES
function ApplyConsistentSplitsToAllDatasets(val_molecules, test_molecules):
    # Load predefined splits
    val_molecules = read_from_file("val_molecules.txt")
    test_molecules = read_from_file("test_molecules.txt")
    
    for each dataset in all_datasets_with_smiles:
        smiles_columns = identify_smiles_columns(dataset)
        
        # Process each row
        for each row in dataset:
            molecules_in_row = extract_molecules_from_columns(row, smiles_columns)
            
            # Apply split priority logic
            if any molecule in molecules_in_row is in test_molecules:
                row.split = "test"
            else if any molecule in molecules_in_row is in val_molecules:
                row.split = "valid"
            else:
                # Random assignment for remaining molecules
                random_value = generate_random_number(0 to 1)
                
                if random_value < train_fraction:
                    row.split = "train"
                else if random_value < train_fraction + val_fraction:
                    row.split = "valid"
                else:
                    row.split = "test"
        
        save_dataset_with_splits(dataset)

# STEP 3: Main execution flow
function main():
    # First perform scaffold split to establish global molecule assignments
    train_molecules, val_molecules, test_molecules = CreateGlobalMoleculeSplits()
    
    # Handle amino acid sequences similarly (not shown)
    # ...
    
    # Apply consistent splits across all remaining datasets with SMILES
    ApplyConsistentSplitsToAllDatasets(val_molecules, test_molecules)
    
    # Handle remaining datasets with random splits
    # ...
\end{codebox}

The same concatenation approach has been implemented for amino-acid sequences, but in this case most datasets are relatively large (at least 200k). After concatenating all sequences, we apply a random train--test--validation split, based on the general idea presented above.

\section{Sampling engine}\label{app:sampling}

\begin{figure}[!h]
    \centering
    \includegraphics[width=0.8\textwidth]{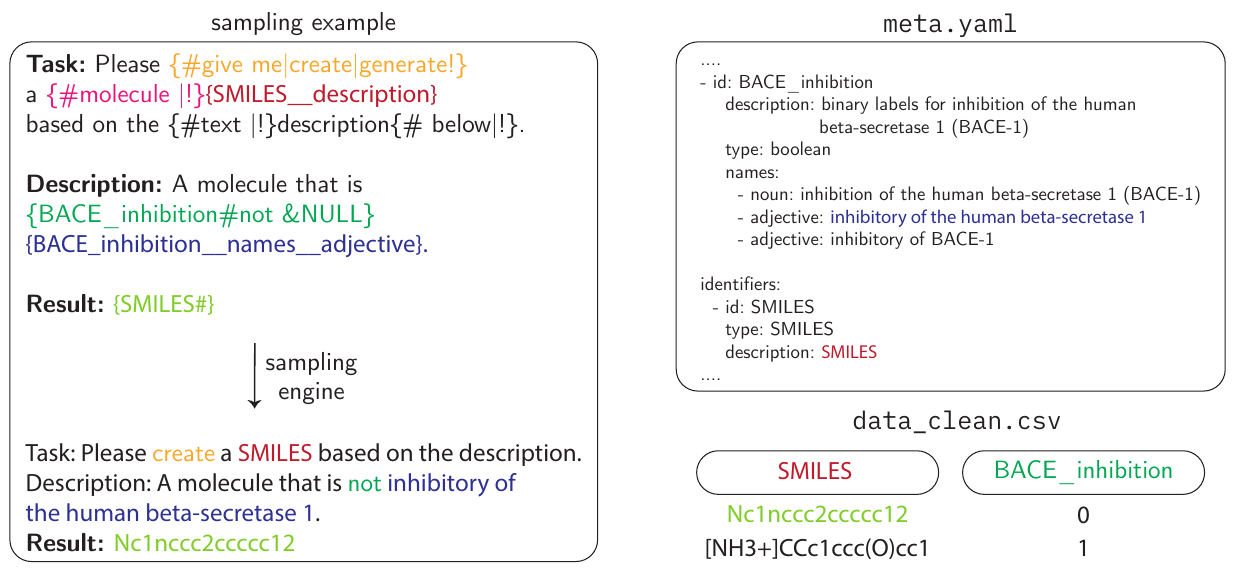}
    \caption{\textbf{Example of how our sampling engine operates.} The sampling depends on two a metadata file, and a raw data file containing all the correct columns as described in the metadata. The colors match what elements of the text templates is replaced in the final text with natural language.}
    \label{fig:sampling_engine}
\end{figure}

Our template sampler consists of more than 800 lines of Python code meant to cover many functionalities. In \Cref{fig:sampling_engine} we show an example of how the sampling engine operates. For the engine to work as intended two files are needed: \texttt{meta.yaml} and \texttt{data\_clean.csv}. The former contains the information about the column names (with specific metadata about semantic types), the text templates, semantic variations of how a property or a representation can be named. The \texttt{meta.yaml} file also contains other metadata such as the URL sources, the citation, the number of points, and a short description of the dataset.

The \texttt{data\_clean.csv} file contains the raw data, with one or more columns for both representations and properties. When sampling, the pipeline extracts the information from this file by pointing to a specific column with the \texttt{\#} (e.g. \texttt{SMILES\#}, BACE\_inhibition\#) symbol. For sampling multiple choice questions the \texttt{\%} is used (e.g. \texttt{SMILES\%} to sample SMILES as options). To indicate the number of MCQ questions, and the type of symbols indicating the different options the following syntax is used: \texttt{\%multiple\_choice\_enum\%2-5\%aA1}. In this example we randomly sample 2 to 5 options. The \texttt{\_\_} component in, for example, \texttt{\{BACE\_inhibition\_\_names\_\_adjective\}} points towards one of the adjectives in the \texttt{names} subfield of the column with the identifier \texttt{BACE\_inhibition}.


In \Cref{tab:templates} we present the five different template types used to generate the \chempile-(M)LIFT datasets. Each example uses special grammar, aforementioned. The pipeline is robust with regards to the representation type, and can include valuable additional information such as units for properties.

\begin{table}[h]
\centering
\caption{\textbf{Template types and examples for each template.} We represent here the five template types used to create the LIFT and (M)LIFT datasets.}
\resizebox{0.7\textwidth}{!}{%
\label{tab:templates}
\begin{tabular}{>{\raggedright\arraybackslash}p{0.2\textwidth} >{\raggedright\arraybackslash}p{0.8\textwidth}}
\toprule
\textbf{Template type} & \textbf{Template} \\
\midrule
& \\[-1.3em]
\textbf{Completion (generative)} & 
The \{\#CIF|CIF file|CIF card!\} of the material with \{\#chemical formula|composition|reduced formula!\} \{formula\#\}, \{spacegroup\_number\_\_names\_\_noun\} \\
&\{spacegroup\_number\#\} and \{density\_\_names\_\_noun\} \{density\#\} \{density\_\_units\} is \{cif\#\}. \\
& \\[-1.3em]
\midrule
& \\[-1.3em]
\textbf{Completion (predictive)} & 
The \{spacegroup\_\_names\_\_noun\} of the symmetrized version of the \{\#material|compound|solid!\} with the \\
& \{\#CIF|CIF file|CIF card!\} \{cif\#\} is \{spacegroup\#\}. \\
& \\[-1.3em]
\midrule
& \\[-1.3em]
\textbf{Instruction (generative)} & 
Task: \{\#Please design|Design!\} a \{\#crystal structure|material|compound|material structure|structure!\} based on the \{cifstr\_\_names\_\_noun\}. \\
& CIF: \{cifstr\#\} \\
& \{\#Description|Answer!\}: \{description\#\} \\
& \\[-1.3em]
\midrule
& \\[-1.3em]
\textbf{Instruction (predictive)} & 
Task: Please \{\#determine|predict|estimate!\} if the \{\#molecule|compound!\} with the \{SMILES\_\_description\} \{SMILES\#\} is \{MUV-713\_\_names\_\_noun\}. \\
& Result: \{MUV-713\#no\&yes\} \\
& \\[-0.8em]
\midrule
& \\[-0.8em]
\textbf{Instruction (multiple-choice)} & 
\{\#Task|Problem statement!\}: Answer the \{\#multiple choice|multiple-choice|MCQ!\} question. \\
& \{\#Question|Query!\}: What is the \{herg\_central\_at\_1uM\_\_names\_\_noun\} of a \{\#compound|drug!\} with the \{SMILES\_\_description\} \{SMILES\#\}? \\
& Constraint: You must return none, one or more options from \{\%multiple\_choice\_enum\%2-5\%aA1\} without using any \{\#other|additional!\} words. \\
& Options: \\
& \{herg\_central\_at\_1uM\%SMILES\%\} \\
& Answer: \{\%multiple\_choice\_result\}. \{herg\_central\_at\_1uM\#\}\{herg\_central\_at\_1uM\_\_units\}. \\
& \\[-0.8em]
\bottomrule
\end{tabular}}
\end{table}

\section{Embeddings for correlation analysis}\label{app:emb_corr}

The embeddings in \Cref{fig:smiles_iupac_correlations} were generated using OpenAI's \texttt{text-embedding-3-large} model. Analysis of 5,000 molecular pairs revealed that the cosine similarity of IUPAC name embeddings demonstrates a stronger correlation with molecular graph chemical similarity (Pearson correlation coefficient r = 0.722) compared to SMILES embeddings (r = 0.521). The difference of 0.201 was evaluated using Fisher's r-to-z transformation, yielding a z-statistic of 16.7 (p < $10^{-9}$), which describes a statistically significant difference between the two Pearson correlations.

\section{Dataset details}
\subsection{\chempile-Education} \label{sec:app-education}

\begin{figure}[h]
    \centering
    \includegraphics[width=\textwidth]{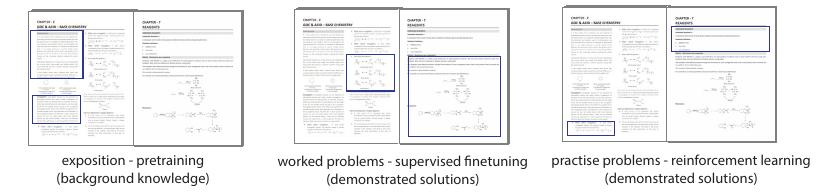}
    \caption{\textbf{\chempile-Education covers different kinds of educational data.} Textbook data contains foundational knowledge, but also worked examples.}
    \label{fig:chempile_smol}
\end{figure}

The general idea pursued by the \chempilelogo-Education dataset is visually presented in \Cref{fig:chempile_smol}. LibreTexts constitute the exposition of the model to diverse background chemistry knowledge, and worked examples. This is further reinforced by lectures from MIT-OCW and YouTube, where examples are often explained and taught step-by-step. Further, we also collected US Olympiad data that can be used in training modes like finetuning or reinforcement learning.

\subsubsection{Sources}

\paragraph{LibreTexts Chemistry}

We systematically extracted and processed the primary textual content from LibreTexts Chemistry HTML documents using a custom Python pipeline---utilizing the Beautiful Soup library for DOM (Document Object Model) parsing. Non-content elements, including navigation menus, scripts, embedded media, acknowledgments, and references, were programmatically removed to isolate chemically relevant educational material. This automated extraction process covered all HTML files in the LibreTexts Chemistry repository (accessed 2025-04-21), generating a structured corpus for subsequent natural language processing analysis. The final dataset contains 114,288,417 tokens across partitioned subsets: training (102,922,903 tokens), validation (5,694,288 tokens), and test sets (5,671,224 tokens).

\paragraph{US Olympiad data}

We manually extracted US Olympiad papers as PDFs from 2003 to 2024 as provided by the American Chemical Society. We used the \texttt{Gemini 2.0 Flash Thinking Experimental 01-21} model and its large context window. Two PDF files were provided to the model: the problem file and the solution file for each year. Based on the problem index, a JSON file was generated with the necessary metadata, question-answer pairs, and answer options. The dataset was then filtered to include only problem solutions that exceeded 250 characters. The rate of success has been evaluated manually on 50 extracted examples. While we do not observe any mismatch between the question and answer pairs, a few minor mismatches are present (e.g. $\Delta$ replaced by $A$).

\paragraph{MIT OpenCourseWare transcripts} To download the data from the MIT OCW we made use of the platform's topic-based search (selecting biology, chemistry, chemical engineering and physics), which allowed us to identify the relevant URL structure for downloading the relevant document. We also provide the course name and the links used to download the course contents.

\paragraph{YouTube transcripts} We find a list of YouTube videos by querying YouTube on a list of LLM-generated keywords. Then, the list of videos is filtered by their license (only the videos labeled as \textit{Creative Commons reuse allowed} were selected). This criterion was achieved by filtering for videos containing \textcolor{blue}{EgIwAQ\%3D\%3D} in the HTML code of respective pages. We then use \texttt{gpt-4.1} to rewrite the raw transcripts into lecture-like content. The advantage of rewriting lies in the inherent gaps in scientific transcripts (i.e., sometimes scientific terms are incorrectly transcribed), which the LLM can fill.

We used the \texttt{gpt-4.1} model to rewrite raw YouTube transcripts into lecture-like content. All transcripts in foreign languages (e.g. Hindi) have also been translated into English. The prompt used to achieve this is given in the snippet below:

\newtcolorbox{promptbox}{
    colback=blue!5!white, 
    colframe=blue!20!black, 
    title=LLM Prompt, 
    fonttitle=\bfseries, 
    breakable, 
    left=2mm, right=2mm, top=2mm, bottom=2mm,
}

\begin{promptbox}
\begin{verbatim}
The following is a transcript of a YouTube video.
Your task is to rewrite the transcript into a lecture format.

Return only the lecture, without any additional text or explanation.
Use the tags [LECTURE] and [/LECTURE] to indicate the start and end of 
the lecture.
The lecture should be structured and easy to follow, feel free to fill 
knowledge gaps.
If the discussion is mathematical, include the equations in LaTeX format.
Same for chemical equations, use the appropriate format.
The lecture should be in English and should not contain any other 
language.

The lecture should be in a single paragraph, without any line breaks.

The transcript is as follows:

{transcript}
\end{verbatim}
\end{promptbox}

\subsection{\chempile-Code dataset}
\paragraph{Distribution of keywords}

\begin{figure}[!h]
    \centering
    \includegraphics[width=0.45\linewidth]{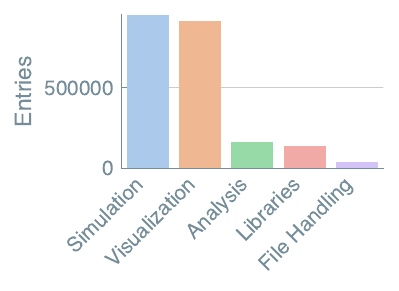}
    \caption{\textbf{Keyword distribution by category}. The plot shows the dataset distribution based on keywords identified in the entry. Here we have considered all the keywords from different categories irrespective of the domain (chemistry, materials science, and biology).}
    \label{fig:keyword-distribution}
\end{figure}

Simulation tools dominate the landscape with the highest number of entries matching simulation tool keywords, indicating their common use in scientific computation across domains. Visualization and Analysis tools follow. Notice that the visualization here is very domain-specific visualization codes (for example, PyMol, VMD, and not matplotlib or plotly). The keywords used for filtering are provided in \Cref{tab:keywords}, and the distribution of the five categories of keywords is shown in \Cref{fig:keyword-distribution}.

\paragraph{Keywords for filtering code}
\Cref{tab:keywords} shows the keywords used to filter and create \chempile-Code dataset.
\begin{table}[h!]
\centering
\caption{\textbf{Overview of Keyword Categories Used for Filtering the Code Dataset.} The table includes all keywords within each predefined list. Note: The \enquote{Keyword Count} reflects the number of distinct terms in each specific list before they are aggregated into a single unique set for regex matching}
\label{tab:keywords}
\resizebox{\textwidth}{!}{%
\begin{tabular}{p{2.5cm}p{2.5cm}p{1cm}p{8cm}}
\toprule
\textbf{Domain} & \textbf{Category} & \textbf{Count} & \textbf{All Keywords} \\
\midrule
Chemistry & Simulation & 10 & GROMACS, LAMMPS, OpenMM, CP2K, Quantum ESPRESSO, NWChem, Psi4, PySCF, ABINIT, Octopus \\
\cmidrule{2-4}
 & Analysis & 7 & MDAnalysis, MDTraj, ChemPy, RDKit, ASE, PySCeS, Open Babel \\
\cmidrule{2-4}
 & Visualization & 6 & VMD, PyMOL, Jmol, Avogadro, Gabedit, RasMol \\
\cmidrule{2-4}
 & File Handling & 5 & Open Babel, Pybel, cclib, Chemfiles, ASE \\
\cmidrule{2-4}
 & Libraries & 8 & RDKit, ChemPy, PySCeS, ASE, OpenFF Toolkit, Chemfiles, Open Babel, CDK \\
\midrule
Materials Science & Simulation & 10 & LAMMPS, Quantum ESPRESSO, ABINIT, SIESTA, Octopus, GPAW, OpenMX, Elk, Elmer FEM, MOOSE \\
\cmidrule{2-4}
 & Analysis & 6 & pymatgen, matminer, phonopy, Matscipy, ASE, MDAnalysis \\
\cmidrule{2-4}
 & Visualization & 6 & OVITO, VMD, ParaView, VTK, VisIt, Mayavi \\
\cmidrule{2-4}
 & File Handling & 5 & ASE, pymatgen, MDAnalysis, MDTraj, Chemfiles \\
\cmidrule{2-4}
 & Libraries & 5 & pymatgen, Matscipy, OpenKIM, pycalphad, ASE \\
\midrule
Biology & Simulation & 6 & NEURON, Brian, COPASI, OpenCOR, Smoldyn, MCell \\
\cmidrule{2-4}
 & Analysis & 9 & BLAST, Bowtie, BWA, Biopython, BioPerl, BioJava, Bioconductor, Galaxy, OpenMS \\
\cmidrule{2-4}
 & Visualization & 4 & Cytoscape, PyMOL, ChimeraX, Napari \\
\cmidrule{2-4}
 & File Handling & 4 & Biopython, pysam, NetCDF, HTSeq \\
\cmidrule{2-4}
 & Libraries & 8 & Biopython, BioPerl, BioJava, scikit-bio, pysam, Bioconductor, Bioconda, Cytoscape \\
\midrule
Other common Quantum Simulations & Software Names & 62 & Gaussian, VASP, ORCA, CASTEP, Amber, Desmond, WIEN2k, NAMD, xTB, MOE, Discovery Studio, BoltzTrap, CHARMM, Wannier90, MOPAC, DMol3, ATK/QuantumATK, Molpro, GROMOS, GAMESS, ADF, TURBOMOLE, Q-Chem, YASARA, Dalton, MacroModel, TINKER, CRYSTAL, FoldX, Jaguar, EPW, RASPA, FHI-aims, FEFF, Hyperchem, GULP, HOOMD-blue, CPMD, CFOUR, FPLO, OpenMolcas, DIRAC, MOLCAS, Yambo, DL\_POLY, PWmat, BerkeleyGW, GPUMD, ESPResSo, Firefly, TeraChem, DFTB+, JDFTx, ACEMD, exciting, FLEUR, QMCPACK, COLUMBUS, deMon2k, TB-LMTO-ASA, ONETEP, CASINO \\
\bottomrule
\end{tabular}
}
\end{table}

\subsection{\chempile-Paper} \label{sec:app-paper}

\subsubsection{Sources}
\paragraph{EuroPMC}

The Europe PMC dataset is a comprehensive, open-access repository of life sciences literature, which includes peer-reviewed full-text research articles, abstracts, and preprints.  Europe PMC houses around 27 million abstracts and 5 million full-text articles. We classified this articles and then filtered out only the abstract and full text articles which are related to chemistry or close to chemistry.

To create a chemistry-specific subset of the extensive Europe PMC dataset, we employed a custom-trained BERT multilabel classifier. This classifier was developed from the ground up using the CAMEL dataset, which provided 20,000 examples for each of the following topics: chemistry, code, math, biology, and physics. The classifier's performance in identifying chemistry-related articles was evaluated on approximately 150 manually annotated entries from the FineWebMath dataset, achieving an $F_1$-score of approximately 0.77 for the chemistry label. We split the document into chunks of 512 tokens, with an overlap of 50 tokens between adjacent chunks, and then took a weighted average of the predictions to determine the topic. We only considered the first five chunks for classifying a document. The abstract-only filtered dataset is over 3 billion tokens.

\paragraph{Specialized preprint servers} To collect chemistry-related articles from specialized preprint servers such as ChemRxiv, BioRxiv and MedRxiv we used the PaperScraper package from \textcite{born2021trends}. All the articles were processed using the Nougat OCR base model from \textcite{blecher2023nougat}. We also collect and distribute metadata such as the license, the publication date, the author list and the title of each preprint.

\paragraph{Arxiv} The Arxiv is a pre-print server initially created for the rapid distribution of physics papers. However, with time, its scope grew to include many other quantitative fields such as materials science, and quantitative biology. Thus, based on the topic keywords \texttt{cond-mat.mtrl-sci} and \texttt{physics.phys-chem}, we extracted the DOIs of chemistry-related articles. We then further used the PaperScraper~\cite{born2021trends} to download the PDF of the respective DOIs.

\paragraph{Materials Safety Data Sheets}
Materials Safety Data Sheets (MSDS) are important resources that disclose the molecular and material safety. We included the tabular form of the MSDS in the language-interfaced tabular data, distinguishing between hazard statements (H) and precautionary statements (P). However, it is often important to describe safety in a more verbose manner. Hence, we converted the PDF version of the MSDS into natural text using the Nougat OCR model from \textcite{blecher2023nougat}

\subsubsection{Post-processing of papers}
\label{sec:clean_papers}
We use a series of regular expression-based filters to remove references (both parenthetical and bracketed citations), figure and schema captions, email addresses. The core function uses year number patterns to identify and truncate citation sections, detecting where reference lists likely begin by finding clusters of publication years, and then cuts the text at the last complete sentence before this section begins. This cleaning process helps to extract the meaningful scientific content from the papers while removing formatting artifacts and reference materials.

\subsection{ChemPile-Reasoning} \label{sec:app-reasoning}

\subsubsection{Sources}
\paragraph{Single-spectra to molecule reasoning traces}

We employed a multiple-choice question framework to generate synthetic reasoning paths for spectral interpretation using \texttt{Claude 3.5 Sonnet} with temperature set to one. Each question presented four candidate molecules alongside a unique spectrum, requiring the model to identify the correct molecular match through structural analysis.

We implemented structured output formatting to facilitate parsing and dataset construction for the single-spectra analysis. Specifically, we instructed the model to encapsulate its reasoning traces between the dedicated tokens [REASONING] and [\textbackslash REASONING], which were subsequently extracted using regular expression pattern matching. Following the approach of \textcite{mirza2024large}, we similarly prompted the model to enclose final answers between [SMILES] and [\textbackslash SMILES] tokens for unambiguous identification.

The validity of predicted SMILES strings was rigorously verified through computational comparison with ground truth structures using \texttt{RDKit}'s molecular object representation. This validation ensured chemical equivalence by comparing molecular graph topologies rather than relying on string matching alone.

\paragraph{Multi-spectra to molecule reasoning traces}

We generated synthetic reasoning paths from spectral data employing a question-based prompting strategy with the \texttt{Deepseek-R1} model, with the temperature set to 0.6. Each prompt included carbon/proton NMR and IR spectra, supplemented with atomic counts, molecular formula, and molecular mass to compensate for the absence of mass spectrometry (MS) data. We evaluated two prompting formats:

\begin{enumerate}
    \item Open-ended questions, requiring free-form generation of the correct SMILES string.
    \item Multiple-choice questions (MCQs), where the model selected the correct answer from structural isomers of the target compound.
\end{enumerate}

For the multi-spectra dataset, we maintained consistency by employing the same SMILES formatting protocol and evaluation methodology as in the single-spectra case. All outputs were parsed using identical regex patterns, with structural validity assessed through RDKit-based comparison against reference structures.

It is important to note that our evaluation criteria were primarily centered on answer accuracy, rather than compliance with formatting requirements. No assessment was conducted regarding the model’s adherence to instructed output formatting guidelines.

The resulting multi-spectra datasets capture questions, step-by-step reasoning traces, final answers, and a boolean correctness label. The open-ended dataset comprises 358,000 tokens, while the MCQ variant contains 946,930 tokens.

\section{Additional embedding visualizations}\label{app:tsne_umap}

\begin{figure}[h]
    \centering
    \includegraphics[width=\textwidth]{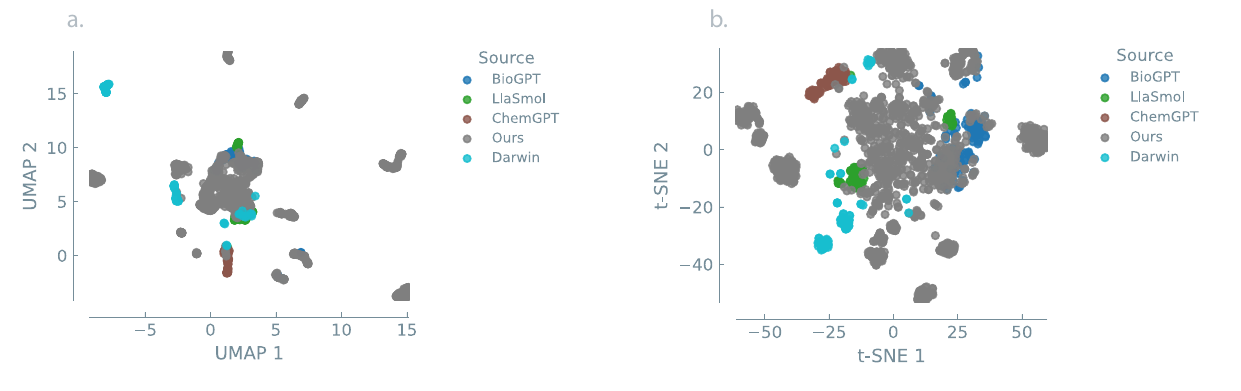}
    \caption{\textbf{Additional embedding dimensionality reduction visualizations for \Cref{fig:scale}b} We use the \texttt{umap-learn} package for UMAP and the \texttt{scikit-learn} package for TSNE. Default settings are used.}
    \label{fig:additional_viz}
\end{figure}

In \Cref{fig:additional_viz} we provide additional visualization for the embeddings in \Cref{fig:scale}b. These reinforce the main idea of \Cref{fig:scale}b, the \chempile is by far the most diverse chemical dataset, capturing a large semantic space.
 
\section{SMILES-IUPAC translation}
Source SMILES sequences are tokenized using Byte-Level Byte Pair Encoding~\cite{sennrich-etal-2016-neural}, while target IUPAC sequences utilize a Unigram tokenizer~\cite{DBLP:conf/acl/Kudo18} with whitespace, punctuation, and digit-level pre-tokenization. The model architecture consists of 8 encoder and 8 decoder layers, an embedding dimension of 1536, 8 attention heads, a 4096-dimensional feed-forward network, and 0.1 dropout. Standard sinusoidal positional encodings are added to scaled token embeddings. The model has been trained for two epochs, with a final training loss of ~0,006 and a validation loss of 0,0089. The model has an accuracy of approximately 91\%. We provide an interface and script for the model on HuggingFace\clickableimage{https://huggingface.co/spaces/AdrianM0/smi2iupac}.

\section{Code snipets}
The following snippets show usage examples of the \chempile. 

\subsection{Using subsets}

\begin{codebox}{\chempilelogo-Caption}{code:chempile-caption}
from datasets import load_dataset

dataset = load_dataset("jablonkagroup/chempile-caption")
print(dataset)
# DatasetDict({
#     train: Dataset({
#         features: ['text', 'image'],
#         num_rows: 90350
#     })
#     validation: Dataset({
#         features: ['text', 'image'],
#         num_rows: 5019
#     })
#     test: Dataset({
#         features: ['text', 'image'],
#         num_rows: 5020
#     })
# })

sample = dataset['train'][0]
print(f"Sample caption: {sample}")
# Sample caption: {'text': '2 drawings and a photograph, as described...', 'image': <PIL...}
\end{codebox}

\subsection{Mixing data}

\begin{codebox}{Obtaining pretraining data-mixes with \chempilelogo}{code:chempile-pretraining}
from datasets import load_dataset, get_dataset_config_names, concatenate_datasets, Dataset
from typing import List

# --- Function to mix data in specified ratios ---
def mix_data_in_ratios(
    grouped_datasets_with_text: List[List[Dataset]],
    ratios: List[float],
    seed: int = 42
) -> Dataset:

    subsampled_data_for_final_mix = []
    for i, group_list in enumerate(grouped_datasets_with_text):
        category_ds = concatenate_datasets(group_list)
        
        num_samples_to_take = int(len(category_ds) * ratios[i])

        if num_samples_to_take > 0:
            selected_subset = category_ds.shuffle(seed=seed).select(range(num_samples_to_take))
            subsampled_data_for_final_mix.append(selected_subset)

    final_mixed_dataset = concatenate_datasets(subsampled_data_for_final_mix)
    return final_mixed_dataset

# --- Main script logic for loading, preparing, and mixing ---
def create_mixed_dataset(category_sources_with_ratios, split):

    dataset_groups_for_mixing = []
    final_ratios_for_mixing = []

    for _, path, ratio in category_sources_with_ratios:
        configs = get_dataset_config_names(path)
        raw_sub_datasets = [load_dataset(path, config, trust_remote_code=True)[split] for config in configs]
        sub_datasets_for_category = []
        for ds in raw_sub_datasets:
            if "text" in ds.column_names and len(ds) > 0:
                sub_datasets_for_category.append(ds.select_columns(["text"]))

        if sub_datasets_for_category:
            dataset_groups_for_mixing.append(sub_datasets_for_category)
            final_ratios_for_mixing.append(ratio)
    
    # Call the mixing function if data is available
    mixed_dataset = mix_data_in_ratios(
        dataset_groups_for_mixing,
        final_ratios_for_mixing,
        seed=42, # for reproducibility
    )
    return mixed_dataset

category_sources_with_ratios = [
    ("education", "jablonkagroup/chempile-education", 1.0),
    ("paper", "jablonkagroup/chempile-paper", 1.0),
    ("code", "jablonkagroup/chempile-code", 0.1)
]
resulting_mixed_dataset = create_mixed_dataset(category_sources_with_ratios, split="train")
\end{codebox}

%% file: sections/datasheets.tex
\clearpage
\section{ChemPile Education Datasheet}
\label{app:edu_datasheet}
\renewcommand{\arraystretch}{1.5}
\begin{longtable}{|p{0.2\textwidth}|p{0.8\textwidth}|}
\hline
\multicolumn{2}{|c|}{\textbf{Dataset Details}} \\
\hline
\endfirsthead
\hline
\endhead
\hline
Purpose of the dataset &  We released ChemPile Education to make Large Language Model training in undergraduate-level chemistry more accessible for the ML community.\\
\hline
Curated by & The dataset was curated by the ChemNLP consortium. \\
\hline
Funded by  & Members of the ChemNLP consortium were supported by different funding sources, which are detailed in the article describing the dataset. A main funding source is the Carl Zeiss Foundation. \\
\hline
Language(s) & English \\
\hline
License & The dataset is released under the Creative Commons (CC) BY-NC-SA 4.0 license. \\
\hline
\multicolumn{2}{|c|}{\textbf{Dataset Structure}} \\
\hline
Data Instances &  The following is an example sample from the dataset. It is part of the \texttt{LibreText\_Chemistry} snapshot and was parsed on \texttt{2025-04-22T23:12:56Z}:

\begin{Verbatim}[breaklines]
{
   "text": "Although not an SI unit, the angstrom (A) is a useful unit of length. It is one ten-billionth of a meter, or 10 -10 m. Why is it a useful unit? The ultimate particles that compose all matter are about 10 -10 m in size, or about 1 Å. This makes the angstrom a natural---though not approved---unit for describing these particles. The angstrom unit is named after Anders Jonas Ångström, a nineteenth-century Swedish physicist. Ångström's research dealt with light being emitted by glowing objects, including the sun. Ångström studied the brightness of the different colors of light that the sun emitted and was able to deduce that the sun is composed of the same kinds of matter that are present on the earth. By extension, we now know that all matter throughout the universe is similar to the matter that exists on our own planet. Anders Jonas Ångstrom, a Swedish physicist, studied the light coming from the sun. His contributions to science were sufficient to have a tiny unit of length named after him, the angstrom, which is one ten-billionth of a meter. Source: Photo of the sun courtesy of NASA's Solar Dynamics Observatory."
}
\end{Verbatim}

\\
\hline
Data Fields & 
- \verb|text (string)|: the text content

\\
\hline
Data Splits & The set is divided into Training, Validation, and Test sets in a ratio of 0.9, 0.1, and 0.1, respectively. \\
\hline
\multicolumn{2}{|c|}{\textbf{Dataset Creation}} \\
\hline
Curation Rationale & With ChemPile Education, we aim to provide the open-source ML community with a clean dataset about chemistry educational resources for pretraining LLMs. \\
\hline
Source Data &  The source data consists of books, course transcripts, and US Olympiad data crawled by the ChemNLP consortium over the 2024-2025 period. \\
\hline
Data processing steps &  The data processing pipeline consists of:

\begin{Verbatim}[breaklines]
- URL filtering
- Text extraction and parsing
- Text filtering and cleaning
\end{Verbatim}
\\
\hline
Annotations &  The dataset does not cover the broad field of chemistry on an undergraduate level.
\\
\hline
Personal and Sensitive Information & Certain author names may persist in the dataset despite text-parsing and cleaning processes. \\
\hline
\multicolumn{2}{|c|}{\textbf{Considerations for Using the Data}} \\
\hline
 Known Limitations & Due to the crawling, some elements might not be correctly filtered, including decorators from the HTML pages or information about the educational contents.  \\
\hline
\end{longtable}

\section{ChemPile LIFT Datasheet}
\label{app:lift_datasheet}
\renewcommand{\arraystretch}{1.5}
\begin{longtable}{|p{0.2\textwidth}|p{0.8\textwidth}|}
\hline
\multicolumn{2}{|c|}{\textbf{Dataset Details}} \\
\hline
\endfirsthead
\hline
\endhead
\hline
Purpose of the dataset &  We released ChemPile LIFT to make Large Language Model training in language-interfaced chemical properties, different nomenclatures, and a diverse set of templates more accessible for the ML community.\\
\hline
Curated by & The dataset was curated by the ChemNLP consortium. \\
\hline
Funded by  & Members of the ChemNLP consortium were supported by different funding sources, which are detailed in the article describing the dataset. A main funding source is the Carl Zeiss Foundation. \\
\hline
Language(s) & English \\
\hline
License & The dataset is released under the Creative Commons (CC) BY-NC-SA 4.0 license. \\
\hline
\multicolumn{2}{|c|}{\textbf{Dataset Structure}} \\
\hline
Data Instances &  The following is an example sample from the dataset. It is part of the \texttt{qm8} snapshot and was parsed on \texttt{2025-04-28T12:24:43Z}:

\begin{Verbatim}[breaklines]
{
    'text': 'The S0 -> S1 transition energy computed using RI-CC2/def2TZVP of the molecule with the SMILES C is 0.433 a. u.'
}
\end{Verbatim}

\\
\hline
Data Fields & 
- \verb|text (string)|: the text content

\\
\hline
Data Splits & The set is divided into Training, Validation, and Test sets in a ratio of 0.9, 0.1, and 0.1, respectively. \\
\hline
\multicolumn{2}{|c|}{\textbf{Dataset Creation}} \\
\hline
Curation Rationale & With ChemPile LIFT, we aim to make accessible a broad range of language-interfaced chemical properties, different nomenclatures, and a diverse set of templates. \\
\hline
Source Data & The source data consists of transforming into text a big amount of the content of several of the most used chemical datasets. \\
\hline
Data processing steps &  The data processing pipeline consists of:

\begin{Verbatim}[breaklines]
- Datasets identification
- Datasets cleaning and pre-processing
- Template gathering
- Template filling
\end{Verbatim}
\\
\hline
Annotations &  The dataset does not cover the broad field of chemistry.
\\
\hline
Personal and Sensitive Information & NA \\
\hline
\multicolumn{2}{|c|}{\textbf{Considerations for Using the Data}} \\
\hline
 Known Limitations & The templates used to contain the data from the datasets are probably not diverse enough.\\
\hline
\end{longtable}

\section{ChemPile Paper Datasheet}
\label{app:paper_datasheet}
\renewcommand{\arraystretch}{1.5}
\begin{longtable}{|p{0.5\textwidth}|p{0.5\textwidth}|}
\hline
\multicolumn{2}{|c|}{\textbf{Dataset Details}} \\
\hline
\endfirsthead
\hline
\endhead
\hline
Purpose of the dataset &  The objective of ChemPile Paper is to make the open-source articles about chemistry more easily accessible for the ML community.\\
\hline
Curated by & The dataset was curated by the ChemNLP consortium. \\
\hline
Funded by  & Members of the ChemNLP consortium were supported by different funding sources, which are detailed in the article describing the dataset. A main funding source is the Carl Zeiss Foundation. \\
\hline
Language(s) & English \\
\hline
License & The dataset is released under the Creative Commons (CC) BY-NC-ND 4.0 license. \\
\hline
\multicolumn{2}{|c|}{\textbf{Dataset Structure}} \\
\hline
Data Instances &  The following is an example sample from the dataset. It is part of the \texttt{euro\_pmc\_chemistry\_papers} snapshot and was parsed on \texttt{2025-05-08T14:08:25Z}:

\begin{Verbatim}[breaklines]
{
    'text': 'Safety and\nBiodistribution of Nanoligomers Targetingthe SARS-CoV-2 Genome for the Treatment of COVID-19 Safety and\nBiodistribution of Nanoligomers Targetingthe SARS-CoV-2 Genome for the Treatment of COVID-19  As the world braces to enter its fourth year of the coronavirusdisease 2019 (COVID-19) pandemic, the need for accessible and effectiveantiviral therapeutics continues to be felt globally. The recent surgeof Omicron variant cases has demonstrated that vaccination and preventionalone cannot quell the spread of highly transmissible variants. Asafe and nontoxic therapeutic with an adaptable design to respondto the emergence of new variants is critical for transitioning tothe treatment of COVID-19 as an endemic disease. Here, we presenta novel compound, called SBCoV202, that specifically and tightly bindsthe translation initiation site of RNA-dependent RNA polymerase withinthe severe acute respiratory syndrome coronavirus 2 (SARS-CoV-2) genome,inhibiting viral replication. SBCoV202 is a Nanoligomer, a moleculethat includes peptide nucleic acid sequences capable of binding viralRNA with single-base-pair specificity to accurately target the viralgenome. The compound has...'
}
\end{Verbatim}

\\
\hline
Data Fields & 
- \verb|text (string)|: the text content

\\
\hline
Data Splits & The set is divided into Training, Validation, and Test sets in a ratio of 0.9, 0.1, and 0.1, respectively. \\
\hline
\multicolumn{2}{|c|}{\textbf{Dataset Creation}} \\
\hline
Curation Rationale & With ChemPile Paper, we aim to provide the open-source ML community with a focused dataset about chemistry research articles resources for pretraining LLMs. \\
\hline
Source Data &  The source data consists of articles collected by the ChemNLP consortium over the 2022-2025 period. \\
\hline
Data processing steps &  The data processing pipeline consists of:

\begin{Verbatim}[breaklines]
- Training and Evaluating classifier model
- Classifying and filtering articles using classifier
- Cleaning of the text
\end{Verbatim}
\\
\hline
Annotations &  The dataset does not cover the broad field of chemical research.
The dataset is incomplete because it does not contain all the information referring to the broad field of chemical research.
\\
\hline
Personal and Sensitive Information & Certain author names may persist in the dataset despite text-parsing and cleaning processes. \\
\hline
\multicolumn{2}{|c|}{\textbf{Considerations for Using the Data}} \\
\hline
 Known Limitations & 
 Certain author names may persist in the dataset despite text-parsing and cleaning processes. Some of the articles in the dataset might not include chemical research and only be related to chemistry.\\
\hline
\end{longtable}

\section{ChemPile Code Datasheet}
\label{app:code_datasheet}
\renewcommand{\arraystretch}{1.5}
\begin{longtable}{|p{0.2\textwidth}|p{0.8\textwidth}|}
\hline
\multicolumn{2}{|c|}{\textbf{Dataset Details}} \\
\hline
\endfirsthead
\hline
\endhead
\hline
Purpose of the dataset &  We release the ChemPile Code dataset to reunite a good amount of code related to chemistry accessible for the ML-community. \\
\hline
Curated by & The dataset was curated by the ChemNLP consortium. \\
\hline
Funded by  & Members of the ChemNLP consortium were supported by different funding sources, which are detailed in the article describing the dataset. A main funding source is the Carl Zeiss Foundation. \\
\hline
Language(s) & English \\
\hline
License & The dataset is released under the Apache License 2.0. \\
\hline
\multicolumn{2}{|c|}{\textbf{Dataset Structure}} \\
\hline
Data Instances &  The following is an example sample from the dataset. It is part of the \texttt{codeparrot\_github-code-chemistry-python} snapshot and was parsed on \texttt{2025-05-08T16:57:40Z}:

\begin{Verbatim}[breaklines]
{
    'text': '####################################
    ####################################\n# This program is copyright (c) Upinder S. Bhalla, NCBS, 2015.\n# It is licenced under the GPL 2.1 or higher.\n# There is no warranty of any kind. You are welcome to make copies under \n# the provisions of the GPL.\n# This programme illustrates building a panel of multiscale models to\n# test neuronal plasticity in different contexts.\n#################################
    #######################################\ntry:\n    import moogli\nexcept Exception as e:\n    print( "[INFO ] Could not import moogli. Quitting..." )\n    quit()\n\nimport numpy\nimport time\nimport pylab\nimport moose\nfrom moose import neuroml\nfrom PyQt4 import Qt, QtCore, QtGui\nimport matplotlib.pyplot as plt\nimport sys\nimport os\nfrom moose.neuroml.ChannelML import ChannelML\nsys.path.append(\'../../../Demos/util\')\n
    import rdesigneur as rd\n\nPI = 3.14159265359\nuseGssa = True\ncombineSegments = True\n# Pick your favourite cell here.\n#elecFileName = "ca1_minimal.p"\n## Cell morphology from Bannister and Larkman J Neurophys 2015/NeuroMorpho\nelecFileName = "h10.CNG.swc"\n#elecFileName = "CA1.morph.xml"\n#elecFileName = "VHC-neuron.CNG.swc"\nsynSpineList = []\nsynDendList = []\nprobeInterval = 0.1\nprobeAmplitude = 1.0\ntetanusFrequency = 100.0\ntetanusAmplitude = 1000\ntetanusAmplitudeForSpines = 1000\nframeRunTime = 1e-3 # 1 ms\nbaselineTime = 0.05\ntetTime = 0.01\npostTetTime = 0.01\nruntime = baselineTime + tetTime + postTetTime\n\ndef buildRdesigneur():\n    \'\'\'\n    #############################
    #####################################\n    # Here we define which prototypes are to be loaded in to the system.\n    # Each specification has the format\n    # source [localName]\n    # source can be any of\n    # filename.extension,   # Identify type of file by extension, load it.\n    # function(),           # func( name ) builds object of specified name\n    # file.py:function() ,  # load Python file, run function(name) in it.\n    # moose.Classname       # Make obj moose.Classname, assign to name.\n    # path...'
}
\end{Verbatim}

\\
\hline
Data Fields & 
- \verb|text (string)|: the main text content

\\
\hline
Data Splits & The set is divided into Training, Validation, and Test sets in a ratio of 0.9, 0.1, and 0.1, respectively. \\
\hline
\multicolumn{2}{|c|}{\textbf{Dataset Creation}} \\
\hline
Curation Rationale & The objective is to curate a subset from big code datasets, and filter the data related to chemistry libraries. \\
\hline
Source Data & The source data is StarCoder and Codeparrot-Github-code\\
\hline
Data processing steps &  The data processing pipeline consists of:

\begin{Verbatim}[breaklines]
- Curating big code dataset
- Filter them based on keywords
- Simple deduplication based on hashing
\end{Verbatim}
\\
\hline
Annotations &  The dataset does not cover all the information and data relative to the broad field of chemistry and all its coding variables and possibilities.
\\
\hline
Personal and Sensitive Information & Due to incomplete filtering, some code snippets contain information relative to the author/s of the code. \\
\hline
\multicolumn{2}{|c|}{\textbf{Considerations for Using the Data}} \\
\hline
 Known Limitations & The accuracy of the classifier used to select the code relative to chemistry is not perfect. Therefore, the dataset might contain some code that is only slightly related to chemistry. \\
\hline
\end{longtable}

\section{ChemPile Reasoning Datasheet}
\label{app:reasoning_datasheet}
\renewcommand{\arraystretch}{1.5}
\begin{longtable}{|p{0.2\textwidth}|p{0.8\textwidth}|}
\hline
\multicolumn{2}{|c|}{\textbf{Dataset Details}} \\
\hline
\endfirsthead
\hline
\endhead
\hline
Purpose of the dataset &  The purpose of ChemPile Reasoning is to provide the community with a clean, well-curated, open-source, and high-quality resource to enhance the reasoning capabilities in chemistry of LLMs.\\
\hline
Curated by & The dataset was curated by the ChemNLP consortium. \\
\hline
Funded by  & Members of the ChemNLP consortium were supported by different funding sources, which are detailed in the article describing the dataset. A main funding source is the Carl Zeiss Foundation. \\
\hline
Language(s) & English \\
\hline
License & The dataset is released under the Creative Commons (CC) BY-SA license. \\
\hline
\multicolumn{2}{|c|}{\textbf{Dataset Structure}} \\
\hline
Data Instances &  The following is an example sample from the dataset. It is part of the \texttt{} snapshot and was parsed on \texttt{2025-04-22T23:12:56Z}:

\begin{Verbatim}[breaklines]
{
    'text': "\nWhich of the following is the correct molecule that corresponds to the given spectra?\n\nA. CC(N)=C(N)C(C)(C)C B. CC(C)=C(N)C(C)(C)N C. CC(C)=C(N)C(C)(C)C D. CSC(C)(C)C(N)=C(C)C\nThe 1H NMR spectrum is as follows: Deltas 1.19 (s, 9H), 1.52 (s, 6H).\nPlease PLACE YOUR ENTIRE REASONING PATHWAY within [START_REASONING][END_REASONING] tags and the correct SMILES within [START_SMILES][END_SMILES] tags.\\n\\nLet's analyze the 1H NMR spectrum step by step:\\n\\n1. We see two signals in the spectrum:\\n   - delta 1.19 (s, 9H)\\n   - delta 1.52 (s, 6H)\\n\\n2. Both signals are singlets (s), which means they don't couple with any other protons.\\n\\n3. The signal at delta 1.19 integrates for 9 protons. This is likely a tert-butyl group (-C(CH3)3).\\n\\n4. The signal at delta 1.52 integrates for 6 protons. This could be two equivalent methyl groups.\\n\\n5. We don't see any signals for NH2 protons, which would typically appear at higher chemical shifts (around 1-5 ppm, broad singlet). This suggests that if there are any nitrogen atoms, they are likely not primary amines.\\n\\n6. We also don't see any signals for alkene protons, which would typically appear at higher chemical shifts (around 4-6 ppm).\\n\\nNow, let's look at the given options:\\n\\nA. CC(N)=C(N)C(C)(C)C - This has two NH groups and would show signals for these protons.\\nB. CC(C)=C(N)C(C)(C)N - This also has two NH groups.\\nC. CC(C)=C(N)C(C)(C)C - This matches our spectrum. It has a tert-butyl group (9H) and two equivalent methyl groups (6H).\\nD. CSC(C)(C)C(N)=C(C)C - This has a sulfur atom and would likely show different chemical shifts.\\n\\nBased on this analysis, the molecule that best matches the given 1H NMR spectrum is option C: CC(C)=C(N)C(C)(C)C\\n\\nThis molecule has:\\n- A tert-butyl group (-C(CH3)3) accounting for the 9H singlet at delta 1.19\\n- Two equivalent methyl groups accounting for the 6H singlet at delta 1.52\\n- No NH protons\\n- No alkene protons (the double bond is tetrasubstituted)\\n\\n", 
}
\end{Verbatim}

\\
\hline
Data Fields & 
- \verb|text (string)|: the main text content
\\
\hline
Data Splits & The set is divided into Training, Validation, and Test sets in a ratio of 0.9, 0.1, and 0.1, respectively. \\
\hline
\multicolumn{2}{|c|}{\textbf{Dataset Creation}} \\
\hline
Curation Rationale & With ChemPile Education, we aim to enrich the chemical reasoning data in the open-source community. \\
\hline
Source Data &  The source data consists of reasoning traces distilled from the leading models by the ChemNLP consortium over the 2024-2025 period. Additionally, it contains Stack Exchange discussions in the field of Materials, Physics, and Chemistry collected over the 2022-2025 period. \\
\hline
Data processing steps &  The data processing pipeline consists of:

\begin{Verbatim}[breaklines]
- URL filtering
- Text extraction
\end{Verbatim}
\\
\hline
Annotations &  The dataset does not cover the broad field of chemical reasoning, and all the chemistry-related tasks.
\\
\hline
Personal and Sensitive Information & Certain user names may persist in the dataset despite text-parsing and cleaning processes. \\
\hline
\multicolumn{2}{|c|}{\textbf{Considerations for Using the Data}} \\
\hline
 Known Limitations & Due to the crawling, some elements might not be correctly filtered, including decorators from the HTML pages or information about the educational contents.  \\
\hline
\end{longtable}

\section{ChemPile MLIFT Datasheet}
\label{app:mlift_datasheet}
\renewcommand{\arraystretch}{1.5}
\begin{longtable}{|p{0.2\textwidth}|p{0.8\textwidth}|}
\hline
\multicolumn{2}{|c|}{\textbf{Dataset Details}} \\
\hline
\endfirsthead
\hline
\endhead
\hline
Purpose of the dataset &  The purpose of the ChemPile MLIFT dataset is to provide the ML community with a comprehensive dataset with language-interfaced chemical property text, accompanied with an image of the molecule involved. \\
\hline
Curated by & The dataset was curated by the ChemNLP consortium. \\
\hline
Funded by  & Members of the ChemNLP consortium were supported by different funding sources, which are detailed in the article describing the dataset. A main funding source is the Carl Zeiss Foundation. \\
\hline
Language(s) & English \\
\hline
License & The dataset is released under the Creative Commons (CC) BY-NC-SA 4.0 license. \\
\hline
\multicolumn{2}{|c|}{\textbf{Dataset Structure}} \\
\hline
Data Instances &  The following is an example sample from the dataset. It is part of the \texttt{} snapshot and was parsed on \texttt{2025-04-22T08:22:19Z}:

\begin{Verbatim}[breaklines]
{
    'SMILES': 'Cc1ccccc1-c1ccc2nc(N)c(C[C@@H](C)C(=O)N[C@@H]3CCOC
    (C)(C)C3)cc2c1', 
    'pIC50': 9.1549015, 
    'BACE_inhibition': 1, 
    'IMAGE': <PIL.PngImagePlugin.PngImageFile image mode=RGB size=300x300 at 0x15481C082A50>, 
    'SELFIES': '[C][C][=C][C][=C][C][=C][Ring1][=Branch1][C][=C][C]
    [=C][N][=C][Branch1][C][N][C][Branch2][Ring1]
    [=Branch2][C][C@@H1][Branch1][C][C][C][=Branch1][C][=O]
    [N][C@@H1][C][C][O][C][Branch1][C][C][Branch1][C][C][C]
    [Ring1][Branch2][=C][C][Ring2][Ring1][Branch1][=C]
    [Ring2][Ring1][=Branch2]', 
    'InChIKey': 'QMSHBBGXSXAGOO-XMSQKQJNSA-N', 
    'IUPAC': '(2R)-3-[2-azanyl-6-(2-methylphenyl)quinolin-3-yl]-
    N-[(4R)-2,2-dimethyloxan-4-yl]-2-methyl-propanamide', 
    'template_original': 'The {#compound|chemical!} with the {SMILES__description} of {SMILES#} {#shows|exhibits|displays!} {BACE_inhibition#no &NULL}{BACE_inhibition__names__noun}.', 
    'template': 'The compound with the SMILES of Cc1ccccc1-c1ccc2nc
    (N)c(C[C@@H](C)C(=O)N[C@@H]3CCOC(C)(C)C3)cc2c1 exhibits inhibition of the human beta-secretase 1 (BACE-1).'}
\end{Verbatim}

\\
\hline
Data Fields & 
- \verb|SMILES (string)|: SMILES representation of the molecule
- \verb|property (float)|: the value of the property relative to the molecule
- \verb|IMAGE (PIL object)|: image of the molecule involved
- \verb|SELFIES (string)|: SELFIES representation of the molecule
- \verb|InChIKey (string)|: InChIKey representation of the molecule
- \verb|IUPAC (string)|: IUPAC name of the molecule involved
- \verb|template_original (string)|: template to adopt with the different representations
- \verb|template (string)|: template to adopt with the different representations
\\
\hline
Data Splits & The set is divided into Training, Validation, and Test sets in a ratio of 0.9, 0.1, and 0.1, respectively. \\
\hline
\multicolumn{2}{|c|}{\textbf{Dataset Creation}} \\
\hline
Curation Rationale & With ChemPile Education, we aim to provide one of the first Multimodal Language-Interfaced datasets relative to chemistry. \\
\hline
Source Data & The source data consists of transforming a large amount of text content from several of the most used chemical datasets into text, providing an image of the involved molecule for each of the rows. \\
\hline
Data processing steps &  The data processing pipeline consists of:

\begin{Verbatim}[breaklines]
- Datasets identification
- Datasets cleaning and pre-processing
- Template gathering
- Image generation
- Representations generation
\end{Verbatim}
\\
\hline
Annotations &  The dataset does not cover all the information related to the broad field of chemistry.
\\
\hline
Personal and Sensitive Information & NA \\
\hline
\multicolumn{2}{|c|}{\textbf{Considerations for Using the Data}} \\
\hline
 Known Limitations & The templates used to contain the data from the datasets are probably not diverse enough.\\
\hline
\end{longtable}

\section{ChemPile Caption Datasheet}
\label{app:caption_datasheet}
\renewcommand{\arraystretch}{1.5}
\begin{longtable}{|p{0.2\textwidth}|p{0.8\textwidth}|}
\hline
\multicolumn{2}{|c|}{\textbf{Dataset Details}} \\
\hline
\endfirsthead
\hline
\endhead
\hline
Purpose of the dataset &  We released ChemPile Caption to make multimodal Large Language Model training in undergraduate-level chemistry more accessible for the ML community.\\
\hline
Curated by & The dataset was curated by the ChemNLP consortium. \\
\hline
Funded by  & Members of the ChemNLP consortium were supported by different funding sources, which are detailed in the article describing the dataset. A main funding source is the Carl Zeiss Foundation. \\
\hline
Language(s) & English \\
\hline
License & The dataset is released under the Creative Commons (CC) BY-NC-SA 4.0 license. \\
\hline
\multicolumn{2}{|c|}{\textbf{Dataset Structure}} \\
\hline
Data Instances &  The following is an example sample from the dataset. It is part of the \texttt{} snapshot and was parsed on \texttt{2025-05-06T18:33:36Z}:

\begin{Verbatim}[breaklines]
{
   'text': 'Figure \\(\\PageIndex{5}\\): Mild cognitive impairment (MCI) is a state between normal ageing and dementia, where someone’s mind is functioning less well than would be expected for their age. This image is for illustrative purposes only. (Public Domain; Center For Functional Imaging, Lawrence Berkeley National Laboratory. Alzheimer’s Disease Neuroimaging Initiative (ADNI).',
   'image': <PIL.JpegImagePlugin.JpegImageFile image mode=RGB size=804x400 at 0x14F34EFF7EC0>
}
\end{Verbatim}

\\
\hline
Data Fields & 
- \verb|text (string)|: the text content
- \verb|image (PIL object)|: the image containing chemical knowledge

\\
\hline
Data Splits & The set is divided into Training, Validation, and Test sets in a ratio of 0.9, 0.1, and 0.1, respectively. \\
\hline
\multicolumn{2}{|c|}{\textbf{Dataset Creation}} \\
\hline
Curation Rationale & With ChemPile Caption, we aim to release an open-source image dataset with images and text at an undergraduate chemistry level. \\
\hline
Source Data &  The source data consists of pairs of image-caption data crawled from LibreTexts Chemistry by the ChemNLP consortium over the 2024-2025 period. \\
\hline
Data processing steps &  The data processing pipeline consists of:

\begin{Verbatim}[breaklines]
- URL filtering
- Text extraction and parsing
- Image extraction and parsing
- Dataset cleaning
\end{Verbatim}
\\
\hline
Annotations &  The dataset does not cover the broad field of chemistry and is biased towards open-source textbook data.
\\
\hline
Personal and Sensitive Information & NA \\
\hline
\multicolumn{2}{|c|}{\textbf{Considerations for Using the Data}} \\
\hline
 Known Limitations & Due to the crawling, some elements might not be correctly filtered, including decorators from the HTML pages or information about the educational contents.  \\
\hline
\end{longtable}